\documentclass[10pt,twocolumn,letterpaper]{article}
\usepackage[pagenumbers]{cvpr}

\usepackage{tikz}
\usetikzlibrary{arrows.meta}

\usepackage{array}
\usepackage{siunitx}
\usepackage{multirow}
\usepackage{colortbl}

\usepackage{fontawesome5}
\definecolor{bdbest}{HTML}{E76254}
\definecolor{bdsecond}{HTML}{4472C4}
\definecolor{cvprblue}{rgb}{0.21,0.49,0.74}
\usepackage[pagebackref,breaklinks,colorlinks,allcolors=cvprblue]{hyperref}
\def\confName{CVPR}
\def\confYear{2027}
\title{Preserve to Compress: Beyond Low-Dimensional Latents\\
  in One-Step Diffusion Codec}
\author{Chuqin Zhou$^{1}$\thanks{Equal contribution.}\quad
  Xiaoyue Ling$^{1}$\footnotemark[1]\quad Yuhang Wei$^{1}$\quad
  Yunuo Chen$^{1}$\quad Guo Lu$^{1}$\thanks{Corresponding author. Email: \href{mailto:luguo2014@sjtu.edu.cn}{luguo2014@sjtu.edu.cn}}\\[3pt]
  $^{1}$Shanghai Jiaotong University}
\begin{document}
\maketitle
\begin{abstract}
Pretrained latent diffusion models bring powerful generative priors to ultra-low-bitrate image compression, but their latent spaces reflect a reconstruction--generation tradeoff. Low-dimensional latents reduce modeling cost but can discard image details, while higher-dimensional latents improve reconstruction yet make diffusion optimization harder under a fixed computational budget. Generative codecs that adopt these pretrained spaces also inherit this compromise, even though compression requires fidelity to a specific source image. In particular, low-dimensional latent interfaces can discard source details before coding and restrict how much decoded information reaches the image decoder. We propose \emph{Preserving Information for Compression and Synthesis} (PICS), a one-step generative image compression framework that adapts these interfaces to preserve source information. Our Information-Preserving Adaptors (IPAs) expose richer features for coding and retain decoded information throughout reconstruction, while deferred decoder adaptation stabilizes training and helps the image decoder exploit the expanded representations. Experiments on Kodak, DIV2K, and CLIC 2020 show average BD-rate savings of 52.64\% and 39.61\% over StableCodec at matched LPIPS and DISTS, respectively.
\end{abstract}

\section{Introduction}
\label{sec:intro}

Ultra-low-bitrate image compression aims to preserve not only the semantic content but also the fine visual details of an image under extremely limited bit budgets~\cite{Jia_24_CVPR_GLC,Li_25_TIP_MISC,Xue_25_ICCV_DLF}. In this regime, minimizing pixelwise distortion favors averaging over plausible reconstructions and yields blurred textures~\cite{Blau_2018_CVPR_PD}. Perceptual codecs alleviate this problem with perceptual and adversarial objectives that push reconstructions toward natural-image statistics~\cite{agustsson_2019_generative,Mentzer_2020_NeurIPS_HiFiC,Muckley_2023_ICML_MSILLM}, and diffusion-based codecs go further by exploiting pretrained generative priors to synthesize realistic detail~\cite{Careil_2024_ICLR_Perco,Li_25_TCSVT_DiffEIC,Ke_25_ICML_ResULIC}. As these priors grow more powerful, realism becomes easier to attain, and the central challenge shifts toward fidelity: the synthesized details should belong to the source image rather than merely look plausible. Fidelity, in turn, hinges on two complementary questions: \emph{how much source information is preserved when the image is encoded, and how much of the transmitted information actually reaches the reconstructed image?}

\begin{figure}[t]
  \centering
  \includegraphics[width=1\linewidth]{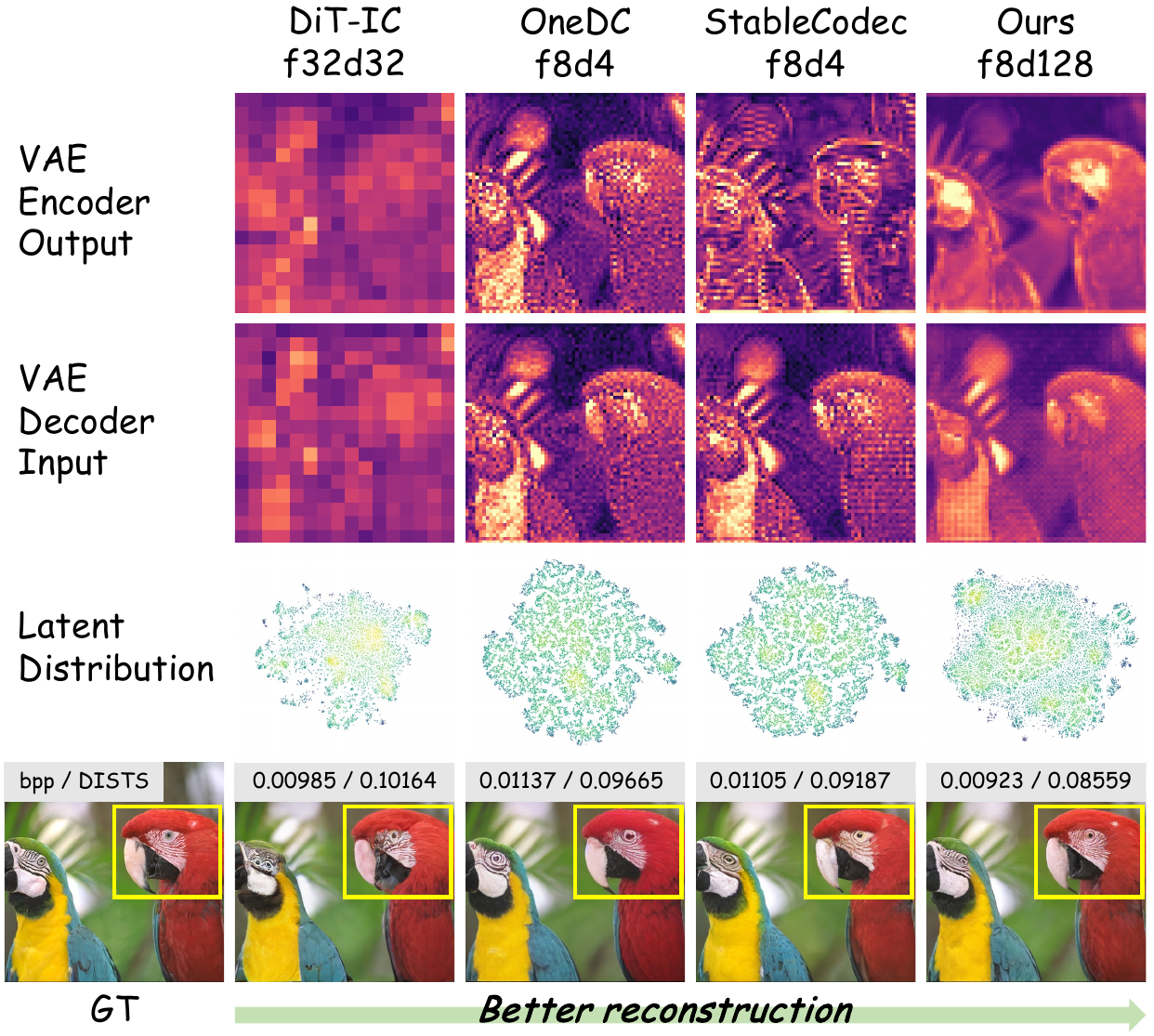}
  \caption{Latent visualization and reconstruction comparison. PICS (Ours) preserves clearer latent structure and finer image details at a lower bitrate. Its more uniform latent distribution may facilitate generative modeling~\cite{Yao_25_CVPR_VAVAE}. Here, $f$ denotes the downsampling factor and $d$ the latent channel count.}
  \label{fig:latent}
  \vspace{-10pt}
\end{figure}

Both questions are shaped by a design choice that diffusion-based codecs rarely revisit: the latent interface. These codecs build on latent diffusion models originally developed for image generation, compressing image latents and using a pretrained diffusion prior to reconstruct the source from the decoded representation~\cite{Relic_24_ECCV_foundation,Zhou_26_TCSVT_DualRep}. Latent diffusion models operate in a low-dimensional latent space learned by an autoencoder to reduce computational cost~\cite{Rombach_2022_CVPR_LDM}, which introduces a reconstruction--generation tradeoff. As analyzed in VA-VAE~\cite{Yao_25_CVPR_VAVAE}, increasing latent dimensionality improves autoencoder reconstruction but makes diffusion optimization harder under a fixed computational budget. Many pretrained generators therefore favor generation, accepting some loss of reconstruction fidelity in exchange for low-dimensional latent interfaces that keep generative modeling tractable. Codecs inherit both these interfaces and the generative prior.

This tradeoff deserves reconsideration in compression, which requires faithful reconstruction of a \emph{specific} source image from the received bitstream. At the encoder, the VAE branch can discard source details when producing low-dimensional latents before rate-constrained coding~\cite{Zhang_25_ICCV_StableCodec,Shi_26_CVPR_DiTIC,Xue_25_NeurIPS_OneDC,Guo_25_NeurIPS_oscar,Huang_2026_arXiv_Gen2IC,Chen_26_ECCV_NeFIC}. Subsequent channel expansion alone cannot recover information discarded within this branch. At the decoder, rich decoded features either enter the denoiser through an expanded input interface but are mapped to low-dimensional output latents~\cite{Zhang_25_ICCV_StableCodec,Xue_25_NeurIPS_OneDC,Sheng_26_CVPR_CADC}, or are mapped to the pretrained latent space before generative refinement~\cite{Shi_26_CVPR_DiTIC,Guo_25_NeurIPS_oscar,Chen_26_ECCV_NeFIC,Chen_26_AAAI_SODEC}. In both cases, the image decoder receives a lower-dimensional representation than the decoded features, limiting use of transmitted information. Such limits can persist with a stronger prior, as plausible details may not match the source.

To preserve source information during both encoding and reconstruction, we propose \emph{Preserving Information for Compression and Synthesis} (PICS), a framework that adapts the latent interfaces inherited from pretrained generators. Our \emph{Information-Preserving Adaptors} (IPAs) expose richer source features for coding at the encoder and keep the decoded information accessible throughout generative reconstruction and image decoding. On the reconstruction side, multi-step diffusion feeds each updated state back into the denoiser and requires compatible input and output representations. One-step decoding sends final features directly to the image decoder, removing this feedback constraint and facilitating adaptation of DiT input and output interfaces.

Preserving richer features through these adapted interfaces is useful only if the image decoder learns to translate them into faithful image detail. Adapting the decoder while the codec and diffusion transformer are still being aligned changes both the reconstruction features and their mapping to pixels simultaneously, complicating optimization. We therefore introduce \emph{deferred decoder adaptation}, which postpones image decoder adaptation until codec--transformer alignment is established. As illustrated in Figure~\ref{fig:latent}, PICS preserves clearer spatial structure at both latent interfaces and recovers finer image details at a lower bitrate. Our contributions are as follows:
\begin{itemize}
  \item We identify the low-dimensional latent interfaces inherited from pretrained generators as a bottleneck for fidelity in diffusion-based compression, and propose PICS, a framework that coordinates IPAs across source encoding and generative reconstruction.
  \item We introduce deferred decoder adaptation, first establishing codec--transformer alignment with a fixed pretrained image decoder, then jointly adapting the reconstruction-side IPAs and the decoder.
  \item Experiments on Kodak, DIV2K, and CLIC 2020 show average BD-rate savings of 52.64\% and 39.61\% over StableCodec at matched LPIPS and DISTS, respectively.
\end{itemize}

\section{Related Work}
\label{sec:related}

\subsection{Learned Image Compression}
Traditional image codecs combine hand-crafted transforms, quantization, and entropy coding~\cite{Sullivan_2012_TCSVT_HEVC,Bellard_2018_BPG,Bross_2021_TCSVT_VVC}. Learned image compression jointly optimizes nonlinear transforms, quantization, and entropy models under a rate--distortion objective~\cite{Theis_17_ICLR_lossy,Balle_2017_ICLR_E2E}, achieving rate--distortion performance beyond strong conventional codecs~\cite{Cheng_2020_CVPR_DGML,He_22_CVPR_ELIC}. This progress comes from both representation learning and entropy modeling. On the representation side, attention, hybrid CNN--Transformer, frequency-domain, state-space, and graph-based architectures improve source feature extraction~\cite{Cheng_2020_CVPR_DGML,Liu_23_CVPR_TCM,Li_24_ICLR_FTIC,Zeng_25_CVPR_MambaIC,Chen_26_CVPR_GLIC}, while differentiable training and vector quantization adapt learned representations to discrete coding~\cite{Agustsson_2017_NeurIPS_SoftToHard,Guo_2021_ICML_SoftThenHard,Feng_23_CVPR_NVTC}. On the entropy side, hyperpriors and autoregressive contexts improve latent probability estimates, and parallel context schedules reduce the sequential cost of decoding~\cite{Balle_2018_ICLR_hyperprior,Minnen_2018_NeurIPS_Joint,He_2021_CVPR_Checkerboard,He_22_CVPR_ELIC}; global references, attention, wavelets, and dictionaries further improve probability prediction~\cite{Qian_2021_ICLR_GlobalReference,Qian_22_ICLR_Entroformer,Fu_24_ECCV_WeConvene,Lu_25_CVPR_DCAE}. Other work improves rate--distortion optimization~\cite{Ali_23_NeurIPS_Correlation,Zhang_25_CVPR_balanced}, adapts representations to individual images~\cite{Yang_2020_NeurIPS_Inference,Kim_24_CVPR_C3,He_24_ICLR_RECOMBINER}, supports variable-rate and progressive coding~\cite{Choi_2019_ICCV_VariableRate,Cui_2021_CVPR_CVR,Jeon_2023_CVPR_CTC}, and enables practical deployment through compact models and compatibility mechanisms~\cite{Wang_2023_ICLR_EVC,Yang_2023_ICCV_Shallow,Tao_2023_ICCV_AdaNIC,Duan_2024_CVPR_BackwardCompatible,Balle_2019_ICLR_IntegerNetworks}.

These codecs typically minimize mean squared error (MSE), which tends to blur fine textures at low bitrates, motivating generative compression~\cite{Blau_2018_CVPR_PD}.

\subsection{Generative Image Compression}
Perceptual image compression balances source fidelity with visual realism under a limited bit budget, a tradeoff formalized by the rate--distortion--perception framework~\cite{Blau_2019_ICML_Rethinking}. Generative codecs use learned image priors to complement transmitted source information, drawing on adversarial generators, generative tokenizers, or diffusion models.

\noindent\textbf{GAN- and tokenizer-based codecs.}
GAN-based codecs combine pixelwise distortion with adversarial and perceptual objectives to improve realism~\cite{agustsson_2019_generative,Mentzer_2020_NeurIPS_HiFiC}. MS-ILLM~\cite{Muckley_2023_ICML_MSILLM} improves statistical fidelity through structured local discriminator supervision, while MRIC~\cite{Agustsson_2023_CVPR_MRIC} offers controllable realism from a shared compressed representation. Tokenizer-based approaches exploit semantically structured generative latents to improve perceptual quality at very low bitrates~\cite{Esser_21_CVPR_VQGAN,Jia_24_CVPR_GLC}. To improve source fidelity, HybridFlow~\cite{Lu_24_MM_Hybridflow} complements discrete codebook tokens with continuous latent features, while DLF~\cite{Xue_25_ICCV_DLF} jointly learns compact semantic tokens and separately coded detail latents to preserve image-specific details beyond the codebook.

\noindent\textbf{Diffusion-based codecs.}
Training-free methods repurpose pretrained diffusion priors for compression through reverse-channel coding, as in DiffC~\cite{Theis_2022_arxiv_DiffC,Vonderfecht_25_ICLR_DiffC}, or noise-codebook selection, as in DDCM and its variants~\cite{Ohayon_25_ICML_DDCM,Vaisman_26_ICLR_TurboDDCM,Zhang_26_IJCAI_TCDDCM}. Avoiding task-specific training simplifies model reuse, but their coding mechanisms and generation--reconstruction tradeoffs constrain performance, while iterative inference remains less efficient than dedicated one-step decoding.

Compression-specific training adapts diffusion priors more directly to source reconstruction. CDC~\cite{Yang_2023_NeurIPS_CDC}, HFD~\cite{Hoogeboom_2023_arXiv_HFD}, and DIRAC~\cite{Ghouse_2023_arXiv_DIRAC} explore multi-step conditional decoding and residual augmentation, while PerCo~\cite{Careil_2024_ICLR_Perco} and DiffEIC~\cite{Li_25_TCSVT_DiffEIC} harness pretrained latent diffusion models for realistic reconstruction at ultra-low bitrates.

Building on these advances, one-step codecs reduce denoising cost while strengthening source guidance. StableCodec~\cite{Zhang_25_ICCV_StableCodec} couples SD-Turbo~\cite{Sauer_2024_ECCV_ADD} with a deep latent codec and auxiliary encoding and reconstruction branches, while OneDC~\cite{Xue_25_NeurIPS_OneDC} conditions synthesis on semantically distilled hyperprior features and adopts DMD2-based distribution matching~\cite{Yin_2024_NeurIPS_DMD2} for perceptual fine-tuning. CADC~\cite{Sheng_26_CVPR_CADC} concentrates source information through content-adaptive quantization and auxiliary reconstruction supervision, supplemented by decoder-derived textual guidance. DiT-IC~\cite{Shi_26_CVPR_DiTIC} extends one-step reconstruction to a deeper transformer latent space through frozen-encoder alignment and latent-derived conditioning. Complementary efforts address computational cost, rate flexibility, and source fidelity: AEIC~\cite{Zhang_26_CVPR_AEIC} uses shallow transforms and a channel-pruned image decoder~\cite{Chen_25_CVPR_AdcSR}; OSCAR~\cite{Guo_25_NeurIPS_oscar} supports multiple bitrates with a shared generative model; and SODEC~\cite{Chen_26_AAAI_SODEC} improves source fidelity through explicit guidance.

CoD~\cite{Jia_26_CVPR_CoD} takes a complementary approach by learning compression-specific diffusion priors from scratch, with its pixel variant bypassing the VAE reconstruction bottleneck. CoD-Lite~\cite{Jia_26_ICML_CoDLite} distills this approach into a lightweight one-step pixel codec. PICS instead adapts pretrained latent interfaces to preserve source information across coding and reconstruction. Its Information-Preserving Adaptors expose richer source features for coding and retain decoded information through generative refinement and image decoding.

\subsection{Diffusion Models}
Diffusion models generate data by learning to reverse a gradual noising process through denoising or score estimation~\cite{Ho_2020_NeurIPS_DDPM,Song_2021_ICLR_ScoreSDE}. Latent diffusion~\cite{Rombach_2022_CVPR_LDM} reduces computational cost by moving this process from pixels to a low-dimensional autoencoder latent space. Diffusion Transformers~\cite{Peebles_2023_ICCV_DiT} use scalable transformer backbones for latent denoising, while flow matching~\cite{Lipman_2023_ICLR_FlowMatching} learns continuous transport from noise to data.

To shorten the iterative generation process, DDIM~\cite{Song_2021_ICLR_DDIM} accelerates sampling, while consistency models~\cite{Song_2023_ICML_Consistency}, adversarial distillation~\cite{Sauer_2024_ECCV_ADD}, and DMD2~\cite{Yin_2024_NeurIPS_DMD2} enable few-step or one-step synthesis. In unconditional or text-conditioned generation, the model must synthesize image content from noise without access to source-specific visual details. Compression provides a different setting, where decoded features carry information about the particular image to be reconstructed. PICS uses these features to adapt a pretrained diffusion model directly for one-step reconstruction through reconstruction and adversarial objectives, without distilling a multi-step teacher.

The effectiveness of reconstruction from these decoded features, in turn, depends on how much source information the latent space preserves. Latent design entails a reconstruction--generation tradeoff. Compact latents reduce generative modeling cost but can discard image details, while higher-dimensional latents can improve reconstruction yet make diffusion optimization more demanding under a fixed computational budget. VA-VAE~\cite{Yao_25_CVPR_VAVAE} analyzes this tension and aligns latent representations with pretrained visual features to facilitate diffusion training while preserving reconstruction quality. DC-AE~\cite{Chen_2025_ICLR_DCAE} improves efficiency by increasing spatial compression while maintaining reconstruction quality. RAE~\cite{Zheng_2026_ICLR_RAE} instead pairs pretrained representation encoders with learned decoders to obtain semantically rich, high-dimensional latents, together with diffusion architectures adapted to these spaces. These studies highlight how latent capacity and structure affect both reconstruction and generation, motivating us to revisit the interfaces inherited by compression codecs to better preserve source-specific details.

\section{Method}
\label{sec:method}

\begin{figure*}[t]
  \centering
  \includegraphics[width=\textwidth]{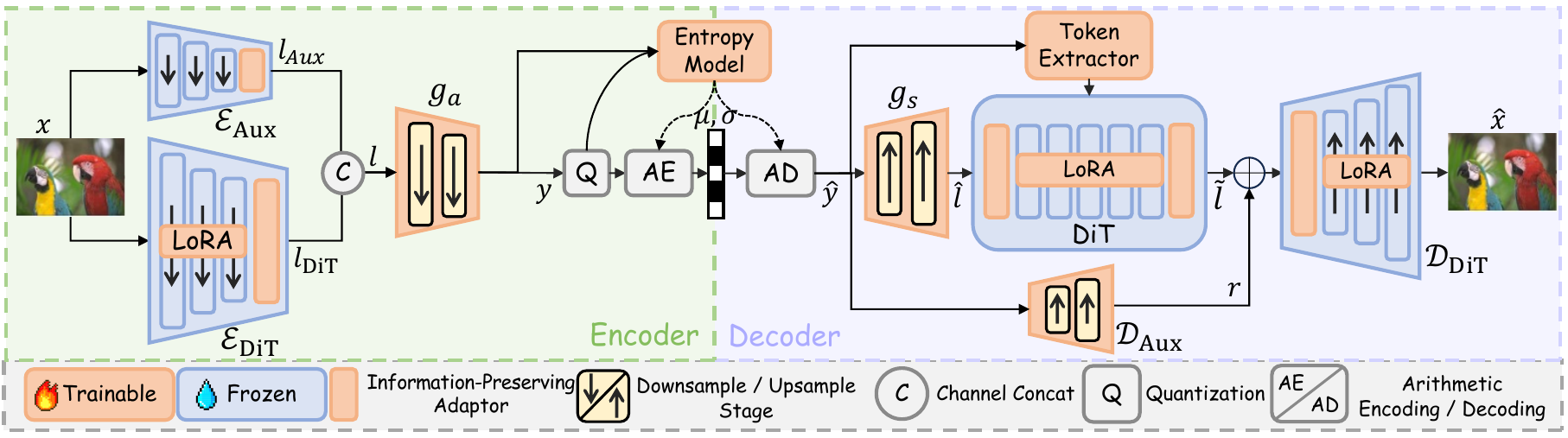}
  \caption{Overview of PICS. Information-Preserving Adaptors (IPAs) expand latent interfaces to preserve source information. VAE and auxiliary features $l_{\mathrm{DiT}}$ and $l_{\mathrm{Aux}}$ are concatenated into $l$ and jointly compressed; $\mu$ and $\sigma$ denote the entropy model's predicted Gaussian mean and standard deviation. The decoded latent $\hat y$ yields synthesis features $\hat l$, visual conditions $\mathbf{C}_{\mathrm{vis}}$, and an auxiliary residual $r$. Conditioned on $\mathbf{C}_{\mathrm{vis}}$, the DiT refines $\hat l$ into $\tilde l$ in one pass; VAE decoding reconstructs $\hat x=\mathcal D_{\mathrm{DiT}}(\tilde l+r)$.}
  \label{fig:architecture}
  \vspace{-10pt}
\end{figure*}

\subsection{Framework Overview}
\label{sec:overview}
As shown in Figure~\ref{fig:architecture}, PICS combines a learned latent codec with a pretrained diffusion transformer (DiT) for one-step generative image compression. IPAs adapt the latent interfaces to preserve richer source information throughout coding and reconstruction. Deferred decoder adaptation stabilizes training and helps the image decoder translate these richer representations into faithful image details.

Given an input image $x$, a VAE encoder $\mathcal E_{\mathrm{DiT}}$ and an auxiliary encoder $\mathcal E_{\mathrm{Aux}}$ extract complementary source features. Their adapted and spatially aligned representations, $l_{\mathrm{DiT}}$ and $l_{\mathrm{Aux}}$, are concatenated and jointly transform-coded:
\begin{equation}
  \begin{aligned}
    l&=\operatorname{Concat}(l_{\mathrm{DiT}},l_{\mathrm{Aux}}),
    &y&=g_a(l),\\
    \hat y&=Q(y),
    &\hat l&=g_s(\hat y).
  \end{aligned}
  \label{eq:coding}
\end{equation}
Here $g_a$ and $g_s$ denote the analysis and synthesis transforms, and $Q$ denotes quantization. Entropy coding uses a hyperprior and a four-step autoregressive entropy model with quadtree partition~\cite{Li_2023_CVPR_DCVCDC}.

At the receiver, a token extractor $T$ derives visual conditions $\mathbf{C}_{\mathrm{vis}}=T(\hat y)$ that guide the adapted DiT. We denote the one-step generative path that refines $\hat l$ under these conditions by $\mathcal G_\theta$. In parallel, an auxiliary decoder predicts a residual $r=\mathcal D_{\mathrm{Aux}}(\hat y)$ directly from the decoded latent. The adapted VAE decoder $\mathcal D_{\mathrm{DiT}}$ reconstructs the image from the sum of the refined features and auxiliary residual:
\begin{equation}
  \tilde l=\mathcal G_\theta(\hat l,\mathbf{C}_{\mathrm{vis}}),\qquad \hat x=\mathcal D_{\mathrm{DiT}}(\tilde l+r).
  \label{eq:reconstruction}
\end{equation}

\subsection{Information-Preserving Adaptors}
\label{sec:interfaces}
The low-dimensional latent interfaces inherited from pretrained generators can restrict both the source features available for coding and the decoded information passed to the image decoder. We introduce IPAs at these interfaces to preserve richer representations throughout coding and reconstruction. IPAs use convolutional or patchwise linear mappings for channel adaptation, with residual feature refinement where appropriate. Their structures are detailed in the supplementary material.

\subsubsection{Preserving Source Features for Coding}
In latent-space compression, the features exposed by the encoder determine which source details are available for coding. The pretrained VAE and DiT share a low-dimensional $c_{\mathrm{DiT}}$-channel latent space, which can restrict the source information available to the rate-constrained analysis transform. Some existing codecs~\cite{Zhang_25_ICCV_StableCodec,Shi_26_CVPR_DiTIC,Xue_25_NeurIPS_OneDC,Guo_25_NeurIPS_oscar} expand the channels of low-dimensional pretrained autoencoder latents. This expansion cannot recover source information already discarded by the autoencoder. We therefore replace the VAE encoder's output mapping with an IPA that directly produces a wider $c_{\mathrm{enc}}$-channel representation in place of the original $c_{\mathrm{DiT}}$-channel latents, making richer source features available for coding. Another IPA prepares complementary features from the frozen auxiliary encoder for fusion. The shared analysis transform $g_a$ then learns which information to retain from the combined representation under the rate--distortion--perception objective.

\subsubsection{Flexible Interfaces for One-Step Reconstruction}
\label{sec:onestep}
Conventional multi-step diffusion repeatedly feeds updated latent states back into the denoiser, requiring denoiser inputs and outputs to share a common latent space. PICS instead performs generative refinement in a single DiT forward pass. The resulting features are combined with the auxiliary residual and passed directly to the image decoder to complete reconstruction. Since no subsequent denoising step consumes the DiT output, the DiT input and output no longer need to share the same channel width. We therefore choose the DiT input and output channel widths independently, adapting the input mapping for decoded synthesis features and the output mapping for image decoding.

The synthesis feature $\hat l$ contains spatial source cues recovered from the bitstream. Mapping it back to the pretrained $c_{\mathrm{DiT}}$-channel latent space before generative refinement can restrict the information accessible to the DiT. We therefore expand the DiT input mapping to $c_{\mathrm{DiT}}^{\mathrm{in}}$ channels with an IPA, allowing the DiT to refine the full synthesis representation under visual conditions $\mathbf{C}_{\mathrm{vis}}$.

A richer input alone does not guarantee that source details remain available to the image decoder. To avoid mapping the refined features back into the original low-dimensional latent space, we jointly expand the DiT output mapping and the VAE decoder input to $c_{\mathrm{DiT}}^{\mathrm{out}}$ channels. The generative output $\tilde l$ and the auxiliary residual $r$ decoded from $\hat y$ share this expanded space and are combined before image decoding, allowing reconstruction to draw on both generative refinement and information supplied directly by the coded latent.

\subsection{Single-Step Reconstruction}
\label{sec:single_step}
PICS reconstructs the image with one DiT evaluation conditioned on a fixed timestep $t_0=1000$. For decoded synthesis features $\hat l\in\mathbb R^{c_{\mathrm{DiT}}^{\mathrm{in}}\times h\times w}$, Eq.~\ref{eq:reconstruction} expands as
\begin{equation}
  \tilde l=\mathcal G_\theta(\hat l,\mathbf{C}_{\mathrm{vis}})=P(\hat l)+g\odot\Phi_\theta(\hat l,t_0,\mathbf{C}_{\mathrm{vis}}),
  \label{eq:single_step}
\end{equation}
where $h$ and $w$ denote the latent spatial height and width, respectively, and $\odot$ denotes elementwise multiplication. The mapping $P$ projects $\hat l$ to a $c_{\mathrm{DiT}}^{\mathrm{out}}$-channel reconstruction state and $\Phi_\theta$ predicts a reconstruction correction. Following the variance-guided reconstruction principle of DiT-IC~\cite{Shi_26_CVPR_DiTIC}, an entropy-conditioned gate $g$ modulates the update strength. The refined features $\tilde l$ and auxiliary residual $r$ decoded from $\hat y$ are combined in the same expanded latent space before VAE decoding.

\subsection{Deferred Decoder Adaptation}
\label{sec:decoder_adaptation}
IPAs make richer source features available to the image decoder, but the decoder must learn to interpret the expanded representation to turn these features into faithful image detail. As the final mapping from latent features to pixels, the decoder directly shapes the reconstructed image and constrains the attainable quality. Adapting it while the codec and DiT are still learning changes both the reconstruction features and their mapping to pixels simultaneously, which can complicate alignment. We therefore first establish codec--transformer alignment with a fixed pretrained image decoder, and only then adapt the reconstruction-side IPAs and the decoder together.

We initialize the single-pass reconstruction network from a pretrained multi-step diffusion model and train PICS in three stages. The codec and IPAs are learned directly, while LoRA~\cite{Hu_2022_ICLR_LoRA} adapts the pretrained VAE encoder, DiT, and VAE decoder.

\noindent\textbf{Stage 1: High-rate alignment.}
We establish codec--transformer alignment at a relatively high bitrate. IPAs are introduced at the VAE encoder output and DiT input. We train the codec and these IPAs together with the VAE encoder and DiT LoRA parameters, while keeping the pretrained image decoder fixed. We optimize
\begin{equation}
  \mathcal L_{\mathrm{S1}}=\lambda R+\mathcal L_{\mathrm{rec}},
  \label{eq:alignment_objective}
\end{equation}
where $R$ is the estimated rate of the main and hyperprior latents, and $\lambda$ controls the rate penalty. The reconstruction loss combines four complementary terms:
\begin{equation}
  \mathcal L_{\mathrm{rec}}=\gamma_1\mathcal L_{\mathrm{pix}}+\gamma_2\mathcal L_{\mathrm{per}}+\gamma_3\mathcal L_{\mathrm{sem}}+\gamma_4\mathcal L_{\mathrm{str}},
  \label{eq:reconstruction_loss}
\end{equation}
where $\gamma_1,\ldots,\gamma_4$ balance pixelwise, perceptual, semantic, and structural supervision. These terms are instantiated as MSE, LPIPS~\cite{Zhang_2018_CVPR_LPIPS}, a CLIP-based feature loss~\cite{Radford_21_ICML_CLIP}, and an edge-aware DISTS loss~\cite{Ding_22_PAMI_DISTS,Li_25_NeurIPS_D3SR}, respectively.

\noindent\textbf{Stage 2: Deferred decoder adaptation.}
Once alignment is established, we introduce the DiT output and VAE decoder input IPAs and train these IPAs together with the VAE decoder LoRA. Source encoding, entropy modeling, and the DiT backbone, including its LoRA parameters, remain fixed, so adaptation focuses on learning how the expanded features map to pixels. Since the rate is independent of the parameters updated in this stage, we optimize only $\mathcal L_{\mathrm{S2}} = \mathcal L_{\mathrm{rec}}$.

\noindent\textbf{Stage 3: Low-rate joint fine-tuning.}
With the expanded reconstruction path in place, we jointly fine-tune the codec, IPAs, and LoRA parameters of the VAE encoder, DiT, and VAE decoder toward the target low bitrates. Rate allocation and generative reconstruction can thus adjust to each other under a tighter bit budget. We additionally introduce adversarial supervision~\cite{Goodfellow_2014_NeurIPS_GAN} and optimize
\begin{equation}
  \mathcal L_{\mathrm{S3}}=\lambda R+\mathcal L_{\mathrm{rec}}
                 +\beta\mathcal L_{\mathrm{adv}},
  \label{eq:objective}
\end{equation}
where $\mathcal L_{\mathrm{adv}}$ encourages realistic image statistics and $\beta$ controls its weight.

\section{Experiments}
\label{sec:experiments}

\begin{figure*}[t]
  \centering
  \includegraphics[width=\textwidth]{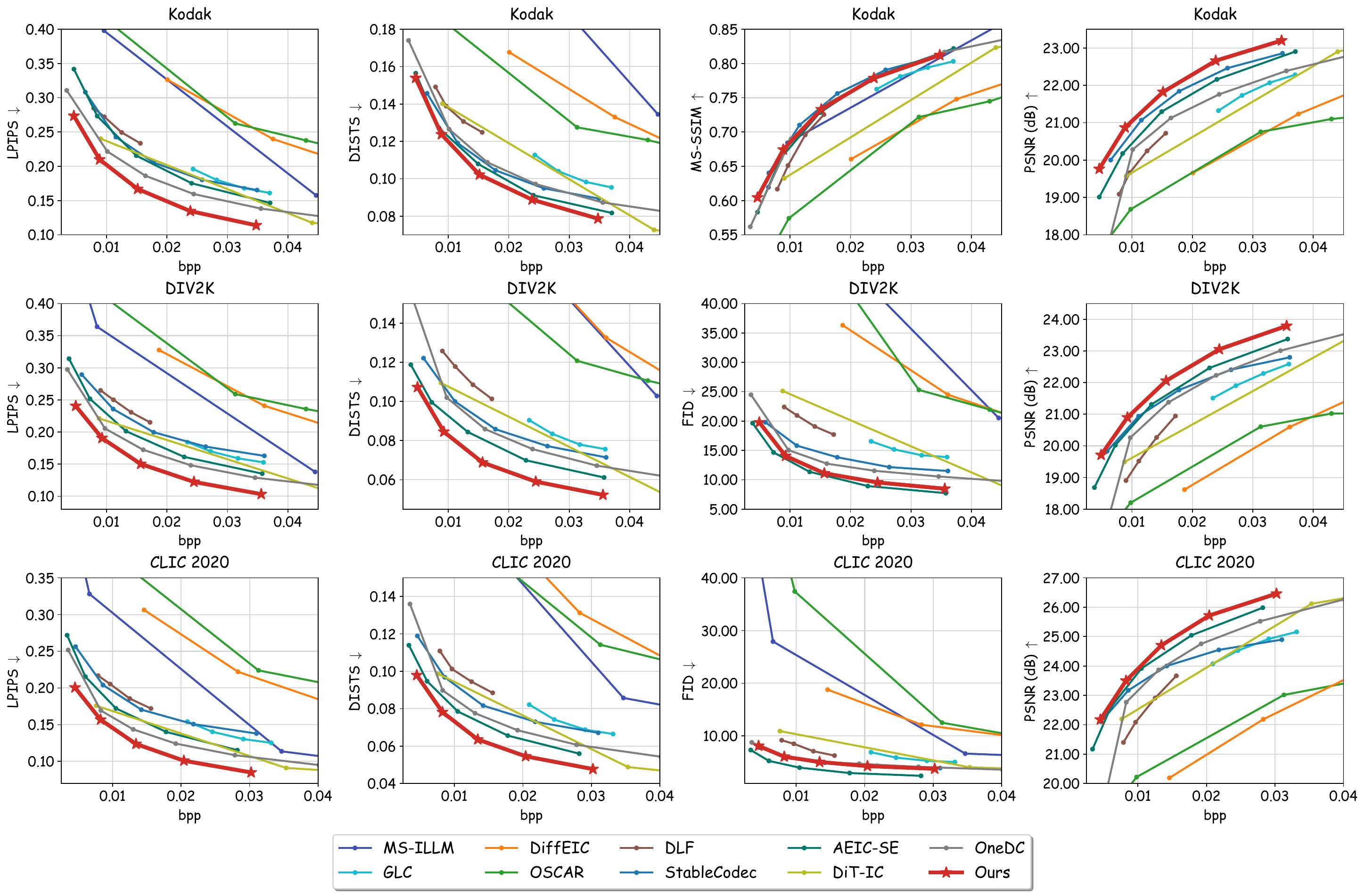}
  \caption{Rate--distortion--perception comparison on Kodak, DIV2K, and CLIC 2020. Lower ($\downarrow$) or higher ($\uparrow$) is better.}
  \label{fig:rate_quality}
\end{figure*}

\subsection{Experimental Setup}
\label{sec:setup}
\noindent\textbf{Datasets.}
We train PICS on the training sets of DIV2K~\cite{Agustsson_2017_CVPRW_DIV2K}, Flickr2K~\cite{Lim_2017_CVPRW_EDSR}, and CLIC 2020 Professional~\cite{Toderici_2020_CLIC}, together with the first 10,000 LSDIR images~\cite{Li_2023_CVPRW_LSDIR}. We evaluate on three benchmarks at their native image resolutions. Kodak~\cite{Kodak_1993_Dataset} contains 24 images with a resolution of $768\times512$ pixels. The DIV2K validation set comprises 100 images at 2K resolution, while the CLIC 2020 test set contains 428 images at 2K resolution.

\noindent\textbf{Implementation details.}
We train PICS with AdamW on four NVIDIA GeForce RTX 5090 GPUs using a total batch size of 8. The diffusion transformer and VAE are both initialized from Lumina-Image 2.0~\cite{Qin_2025_ICCV_LuminaImage2} and adapted using LoRA and IPAs. The original latent width is $c_{\mathrm{DiT}}=16$, and the adapted widths are $c_{\mathrm{enc}}=128$, $c_{\mathrm{DiT}}^{\mathrm{in}}=320$, and $c_{\mathrm{DiT}}^{\mathrm{out}}=128$. Training follows the three-stage schedule in Sec.~\ref{sec:decoder_adaptation}, with $512\times512$ crops. Stages 1 and 2 run for 100K and 40K steps with a fixed learning rate of $10^{-4}$, respectively. Stage 3 runs for 20K steps, starting with an initial learning rate of $5\times10^{-5}$ that is progressively scaled by $0.4$, $0.5$, and $0.1$ at 5K, 10K, and 15K steps, respectively. We set $\lambda=2.5$ in Stage 1 and use $\lambda\in\{4,8,16,32,64\}$ in Stage 3 to obtain different rate operating points. The discriminator uses an initial learning rate of $2\times10^{-5}$. Following~\cite{Zhang_25_ICCV_StableCodec,Zhou_26_TCSVT_DualRep}, high-resolution inference processes overlapping latent tiles for $1024\times1024$ image regions in parallel and blends their outputs using normalized Gaussian weights.

\noindent\textbf{Evaluation metrics.}
We compute bitrate in bits per pixel (bpp) by dividing the actual encoded file size in bits by the number of pixels in the source image. PSNR and MS-SSIM~\cite{Wang_03_ACSSC_MSSSIM} quantify reconstruction fidelity, while LPIPS~\cite{Zhang_2018_CVPR_LPIPS} and DISTS~\cite{Ding_22_PAMI_DISTS} measure perceptual similarity to the source. On DIV2K and CLIC 2020, we also assess distribution-level realism with FID~\cite{Heusel_2017_NeurIPS_FID}, computed on overlapping $256\times256$ RGB patches following HiFiC~\cite{Mentzer_2020_NeurIPS_HiFiC}; see supplementary for details. Negative BD-rates indicate bitrate savings at matched quality; see the appendix for computation details.

\noindent\textbf{Compared methods.}
We compare PICS with the GAN-based codec MS-ILLM~\cite{Muckley_2023_ICML_MSILLM} and the tokenizer-based codecs GLC~\cite{Jia_24_CVPR_GLC} and DLF~\cite{Xue_25_ICCV_DLF}. The diffusion-based baselines include the multi-step codec DiffEIC~\cite{Li_25_TCSVT_DiffEIC} and the one-step codecs OSCAR~\cite{Guo_25_NeurIPS_oscar}, StableCodec~\cite{Zhang_25_ICCV_StableCodec}, AEIC-SE~\cite{Zhang_26_CVPR_AEIC}, DiT-IC~\cite{Shi_26_CVPR_DiTIC}, and OneDC~\cite{Xue_25_NeurIPS_OneDC}.

\begin{figure*}[t]
  \centering
  \includegraphics[width=\textwidth]{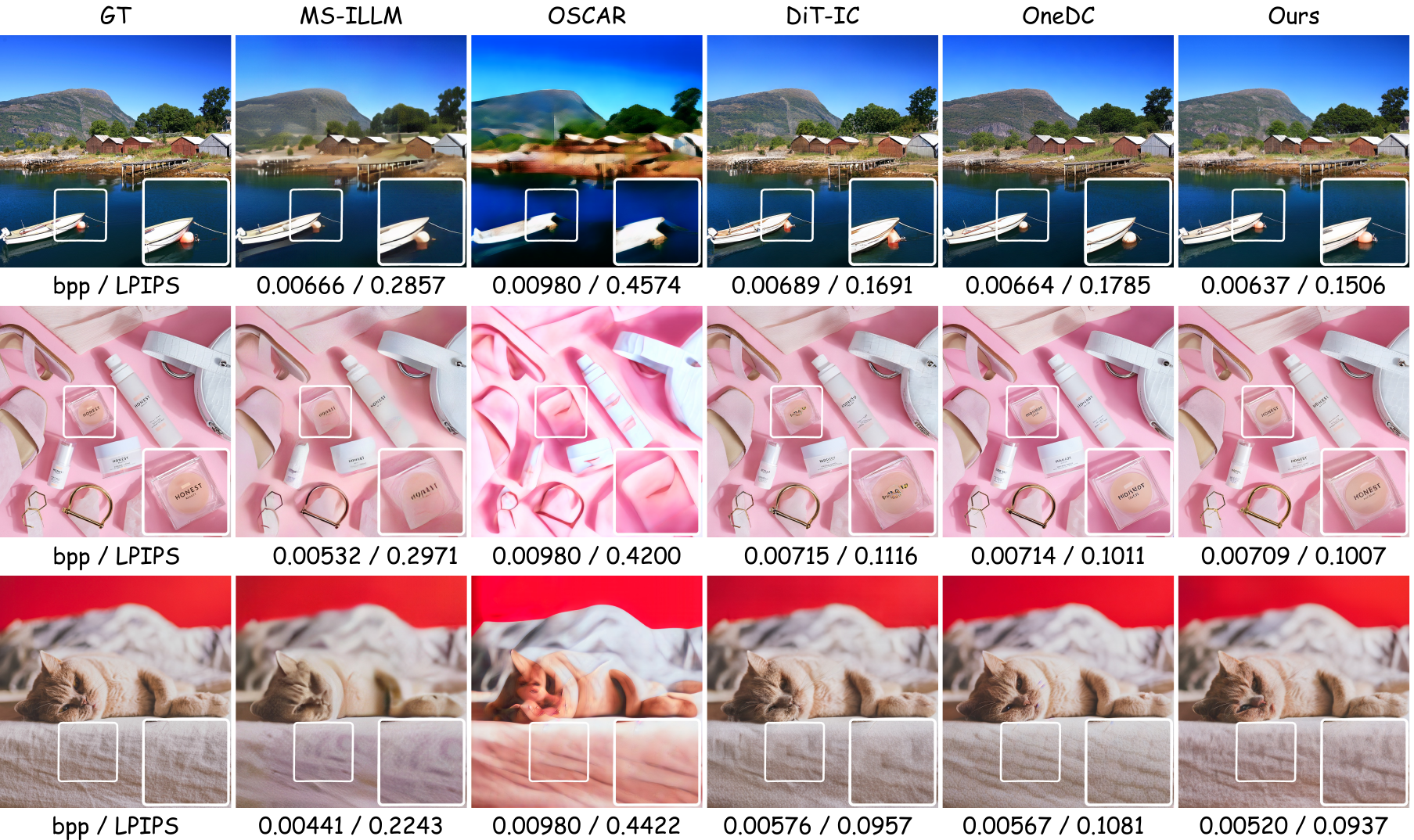}
  \caption{Visual comparison at ultra-low bitrates. Labels report full-image bpp / LPIPS (lower is better); crops are for visualization.}
  \label{fig:qualitative}
\end{figure*}

\subsection{Comparison with State-of-the-Art Methods}
\label{sec:main_results}
\noindent\textbf{Quantitative comparison.}
Figure~\ref{fig:rate_quality} shows that PICS improves perceptual coding efficiency across all three benchmarks. Relative to StableCodec, PICS achieves LPIPS and DISTS BD-rate savings of 55.42\% and 50.56\% on DIV2K, and 54.49\% and 50.34\% on CLIC 2020, respectively. The corresponding averages across the three datasets reach 52.64\% and 39.61\%. Against DiT-IC and OneDC, PICS achieves DISTS BD-rate savings of 38.77\% and 47.41\% on DIV2K, and 32.78\% and 39.49\% on CLIC 2020, respectively. These perceptual gains are accompanied by higher PSNR than StableCodec, DiT-IC, and OneDC on all three datasets, supporting information preservation for both perceptual quality and source fidelity. Detailed BD-rate results and computation procedures are provided in the supplementary material.

\noindent\textbf{Qualitative comparison.}
Figure~\ref{fig:qualitative} compares local structure, lettering, and texture at ultra-low bitrates. In the harbor scene, PICS preserves the rounded buoy beside the boat and its red and white coloring. MS-ILLM blurs this small object, while the other compared generative methods alter its shape or color. In the cosmetics scene, PICS retains more legible lettering on the compact. MS-ILLM and OSCAR largely lose the text, whereas DiT-IC and OneDC distort the character shapes. In the cat example, PICS reconstructs more realistic fabric texture beneath the animal, with visible folds and local texture variations. These examples highlight the importance of retaining source-specific details, since a plausible overall appearance can coexist with distorted object contours or lettering. In all three cases, PICS achieves the lowest LPIPS among the displayed methods while using fewer bits than DiT-IC and OneDC, consistent with the quantitative comparison.

\begin{table}[t]
  \centering
  \small
  \caption{BD-rate (\%) ablation on Kodak using PSNR, LPIPS, and DISTS. Variants A--E progressively add components. All BD-rates are relative to Ours; $\Delta$Params (M) reports additional parameters relative to A. In./Out.\ IPA: DiT input/output IPA; Enc./Dec.: VAE encoder/decoder.}
  \label{tab:interfaces}
  \setlength{\tabcolsep}{1.5pt}
  \renewcommand{\arraystretch}{1.12}
  \sisetup{table-format=2.2, detect-weight=true, mode=text}
  \begin{tabular}{@{}l S[table-format=1.2] S S S@{}}
    \toprule
    \multicolumn{1}{c}{Variant} & {$\Delta$Params} & {PSNR$\downarrow$} & {LPIPS$\downarrow$} & {DISTS$\downarrow$} \\
    \midrule
    \textbf{A}: Base               & 0.00 & 26.71 & 37.42 & 39.34 \\
    \textbf{B}: A + In.\ IPA       & 2.80 & 20.85 & 25.65 & 26.17 \\
    \textbf{C}: B + Enc.\ IPA      & 3.86 & 19.50 & 22.46 & 25.39 \\
    \textbf{D}: C + Token extractor     & 5.74 & 10.15 & 11.53 & 12.97 \\
    \textbf{E}: D + Dec.\ LoRA     & 6.77 &  1.64 &  4.96 &  8.65 \\
    \midrule
    \rowcolor{gray!12}
    \textbf{Ours}: E + Dec.\ \& Out.\ IPAs
      & 8.84 & \bfseries 0.00 & \bfseries 0.00 & \bfseries 0.00 \\
    \bottomrule
  \end{tabular}
\end{table}

\subsection{Ablation Studies}
\label{sec:ablation}

\noindent\textbf{Cumulative component ablation.}
Table~\ref{tab:interfaces} reports cumulative ablations on Kodak, with Ours as the BD-rate anchor. The DiT input IPA (B) reduces the LPIPS and DISTS BD-rate penalties from 37.42\% and 39.34\% to 25.65\% and 26.17\%. This supports retaining the full synthesis representation for DiT refinement. The VAE encoder IPA (C) further improves all three metrics, consistent with making richer source features available before coding. The token extractor (D) and VAE decoder LoRA (E) bring additional gains. Crucially, the DiT output and VAE decoder input IPAs improve all three BD-rates beyond decoder LoRA alone, closing the remaining LPIPS and DISTS penalties of 4.96\% and 8.65\% in E. This highlights the complementary role of reconstruction-side IPAs in retaining information through image decoding. Together, these additions require only 8.84M parameters over A, approximately 0.32\% of our 2.795B-parameter model.

\noindent\textbf{Deferred decoder adaptation.}
\label{sec:decoder_results}
Table~\ref{tab:decoder_timing} compares immediate and deferred reconstruction adaptation under the same total training budget, using \emph{Frozen body} (variant D in Table~\ref{tab:interfaces}) as the anchor. Both strategies add the VAE decoder LoRA, VAE decoder input IPA, and DiT output IPA, either from the start or after codec--transformer alignment. Immediate adaptation improves PSNR but worsens LPIPS and DISTS, consistent with the decoder LoRA distortion--perception tradeoff in StableCodec~\cite{Zhang_25_ICCV_StableCodec}. We hypothesize that adapting the diffusion model and decoder before alignment can destabilize the pretrained generative prior by simultaneously changing latent reconstruction and its mapping to pixels. Deferred adaptation performs best across all three metrics, with PSNR, LPIPS, and DISTS BD-rate savings of 9.22\%, 10.34\%, and 11.48\% over Frozen body, respectively, supporting adaptation after alignment.

\begin{table}[t]
  \centering
  \small
  \caption{BD-rate (\%) of reconstruction-adaptation strategies on Kodak under the same total training budget, computed with PSNR, LPIPS, and DISTS as quality metrics. All BD-rates are relative to Frozen body. Adaptation comprises the VAE decoder LoRA, the VAE decoder input IPA, and the DiT output IPA.}
  \label{tab:decoder_timing}
  \setlength{\tabcolsep}{4pt}
  \renewcommand{\arraystretch}{1.12}
  \sisetup{table-format=-2.2, detect-weight=true, mode=text}
  \begin{tabular}{@{}l S S S@{}}
    \toprule
    \multicolumn{1}{c}{Decoder strategy} & {PSNR$\downarrow$} & {LPIPS$\downarrow$} & {DISTS$\downarrow$} \\
    \midrule
    Adapt from the start & -5.74 &   3.67 &   2.68 \\
    \rowcolor{gray!12}
    Deferred adaptation  & \bfseries -9.22 & \bfseries -10.34 & \bfseries -11.48 \\
    \bottomrule
  \end{tabular}
\end{table}

\begin{figure}[t]
  \centering
  \includegraphics[width=\columnwidth]{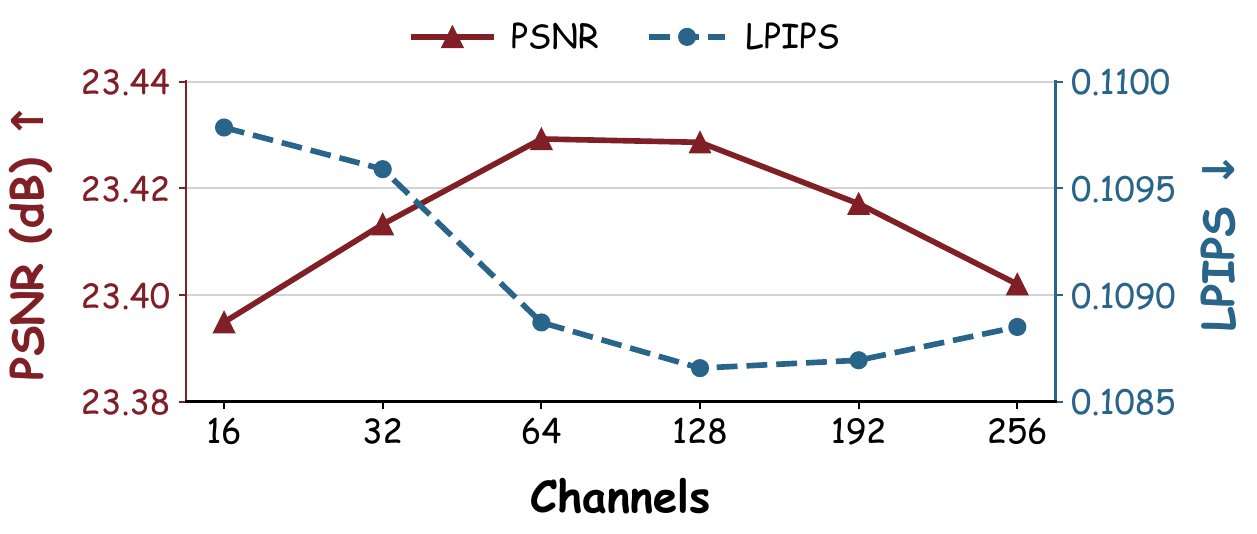}
  \caption{Reconstruction interface width versus PSNR and LPIPS on Kodak using identical bitstreams across widths.}
  \label{fig:output_channels}
\end{figure}

\noindent\textbf{Reconstruction interface width.}
\label{sec:output_width}
During deferred decoder adaptation, we vary $c_{\mathrm{DiT}}^{\mathrm{out}}$ from 16 to 256 and adjust the VAE decoder input and reconstruction mappings accordingly. Source encoding, entropy modeling, and the DiT backbone with its LoRA parameters remain fixed; all widths use identical Kodak bitstreams. Figure~\ref{fig:output_channels} shows that widening the native 16-channel interface improves both metrics: PSNR peaks near 64--128 channels, while LPIPS is lowest at 128. Wider interfaces slightly degrade both, possibly admitting more noise or source-inconsistent features. We therefore use 128 channels.

\noindent\textbf{Prompt conditioning.}
The visual conditions $\mathbf{C}_{\mathrm{vis}}$ guide reconstruction with information derived from the decoded latent. Table~\ref{tab:prompt} evaluates tokens alone and their combination with a fixed prompt, using fixed-prompt conditioning as the anchor. Fixed-prompt conditioning and tokens only correspond to variants C and D in Table~\ref{tab:interfaces}, respectively. At approximately 0.0238 bpp, tokens alone achieve the best measured values on all four metrics, improving PSNR by 0.15\,dB and reducing LPIPS and DISTS by 0.0070 and 0.0063 relative to the fixed prompt. This favors source-dependent conditioning at the evaluated operating point; adding the fixed prompt brings no further benefit.

\begin{table}[t]
  \centering
  \small
  \caption{BD-rate (\%) ablation of prompt conditioning on Kodak, computed with PSNR, MS-SSIM, LPIPS, and DISTS as quality metrics. All BD-rates are relative to fixed-prompt conditioning.}
  \label{tab:prompt}
  \setlength{\tabcolsep}{3.5pt}
  \renewcommand{\arraystretch}{1.12}
  \sisetup{table-format=-1.2, detect-weight=true, mode=text}
  \begin{tabular}{@{}l S S S S@{}}
    \toprule
    \multicolumn{1}{c}{Conditioning} & {PSNR$\downarrow$} & {MS-SSIM$\downarrow$} & {LPIPS$\downarrow$} & {DISTS$\downarrow$} \\
    \midrule
    Prompt + tokens & -4.84 & -3.27 & -6.06 & -8.44 \\
    \rowcolor{gray!12}
    Tokens only     & \bfseries -7.82 & \bfseries -5.16 & \bfseries -8.92 & \bfseries -9.91 \\
    \bottomrule
  \end{tabular}
\end{table}

\subsection{Computational Complexity}
\label{sec:complexity}
Table~\ref{tab:cost} compares perceptual coding efficiency and runtime on Kodak. We evaluate all methods on a single RTX 5090 with batch size 1 and report the mean latency over twenty timed runs per image following five warm-up runs. Timing includes entropy coding but excludes initialization, image I/O, and data transfers. PICS requires 75.77\,ms for encoding and 186.41\,ms for decoding, with a single whole-image transformer evaluation during decoding. Compared with StableCodec, PICS achieves LPIPS and DISTS BD-rate savings of 48.00\% and 17.93\%, respectively, while increasing total latency by 2.45\%, from 255.92\,ms to 262.18\,ms. Additional runtime comparisons at $1024\times1024$ resolution are provided in the appendix.

\begin{table}[t]
  \centering
  \small
  \caption{Perceptual coding efficiency and runtime on Kodak. BD-rates (\%) are relative to StableCodec. Runtime is measured on one RTX 5090 in milliseconds per image.}
  \label{tab:cost}
  \setlength{\tabcolsep}{2pt}
  \renewcommand{\arraystretch}{1.12}
  \sisetup{detect-weight=true, mode=text}
  \begin{tabular}{@{}l
                  S[table-format=-3.2]
                  S[table-format=-3.2]
                  S[table-format=3.2]
                  S[table-format=4.2]@{}}
    \toprule
    \multirow{2}{*}{Method}
      & \multicolumn{2}{c}{BD-rate (\%)}
      & \multicolumn{2}{c}{Runtime (ms)} \\
    \cmidrule(lr){2-3}\cmidrule(l){4-5}
      & {LPIPS$\downarrow$} & {DISTS$\downarrow$} & {Enc.} & {Dec.} \\
    \midrule
    MS-ILLM (ICML'23)~\cite{Muckley_2023_ICML_MSILLM} &   91.09 &  469.19 & 155.04 &  148.45 \\
    GLC (CVPR'24)~\cite{Jia_24_CVPR_GLC} &    9.02 &   57.16 &  41.80 &   65.63 \\
    DLF (ICCV'25)~\cite{Xue_25_ICCV_DLF} &   13.58 &   37.54 &  62.69 &  132.01 \\
    DiffEIC (TCSVT'25)~\cite{Li_25_TCSVT_DiffEIC} &  193.32 &  261.26 & 152.98 & 5206.05 \\
    OSCAR (NeurIPS'25)~\cite{Guo_25_NeurIPS_oscar} &  221.63 &  232.88 &  61.50 &  205.66 \\
    StableCodec (ICCV'25)~\cite{Zhang_25_ICCV_StableCodec} &    0.00 &    0.00 &  99.90 &  156.02 \\
    AEIC-SE (CVPR'26)~\cite{Zhang_26_CVPR_AEIC} &   -5.76 &  -13.00 &  11.47 &   70.23 \\
    DiT-IC (CVPR'26)~\cite{Shi_26_CVPR_DiTIC} &  -33.96 &    7.60 & 101.80 &  165.50 \\
    OneDC (NeurIPS'25)~\cite{Xue_25_NeurIPS_OneDC} &  -34.82 &    3.98 &  80.52 &  182.00 \\
    \midrule
    \rowcolor{gray!12}
    \textbf{Ours} & \bfseries -48.00 & \bfseries -17.93 & 75.77 & 186.41 \\
    \bottomrule
  \end{tabular}
\end{table}

\section{Conclusion}
\label{sec:conclusion}
We presented PICS, a one-step generative image codec that balances tractable generation and faithful reconstruction by adapting pretrained latent interfaces. Information-Preserving Adaptors expose richer source features before rate-constrained coding and retain decoded information through generative refinement and image decoding. Deferred decoder adaptation helps exploit these representations after codec--transformer alignment. Experiments on Kodak, DIV2K, and CLIC 2020 show substantial gains in perceptual coding efficiency while preserving structure and fine details at ultra-low bitrates. These results establish information-preserving latent interfaces as an effective design principle for faithful generative compression.

{
  \small
  \bibliographystyle{ieeenat_fullname}
  \bibliography{main}
}

\clearpage
\appendix
\setcounter{figure}{0}
\renewcommand{\theHfigure}{appendix.\arabic{figure}}
\setcounter{table}{0}
\renewcommand{\theHtable}{appendix.\arabic{table}}
\setcounter{equation}{0}
\renewcommand{\theHequation}{appendix.\arabic{equation}}
\renewcommand{\thefigure}{A\arabic{figure}}
\renewcommand{\thetable}{A\arabic{table}}
\renewcommand{\theequation}{A\arabic{equation}}
\twocolumn[{\centering\Large\bfseries Supplementary Material\par\vspace{1em}}]

\noindent
The main paper introduces Preserving Information for Compression and Synthesis (PICS), a one-step image codec that expands latent interfaces to preserve source detail. We give implementation details and further results below.
\begin{itemize}
  \item Section~\ref{sec:appendix_implementation}. Flow matching, adaptor architecture, visual conditioning, and training settings.
  \item Section~\ref{sec:appendix_protocols}. Latent visualization, FID, and BD-rate.
  \item Section~\ref{sec:appendix_runtime}. Codec latency on $1024\times1024$ images.
  \item Section~\ref{sec:appendix_bd_comparison}. BD-rates on Kodak, DIV2K, and CLIC 2020.
  \item Section~\ref{sec:appendix_baseline_code}. Code repositories for the baseline codecs.
  \item Section~\ref{sec:appendix_visual}. Reconstructed images at ultra-low bitrates.
\end{itemize}

\section{Model Architecture and Training}
\label{sec:appendix_implementation}

\subsection{Flow Matching}
\label{sec:appendix_lumina_flow}
We adopt Lumina-Image 2.0~\cite{Qin_2025_ICCV_LuminaImage2} as the pretrained DiT backbone in PICS. It is trained with flow matching~\cite{Lipman_2023_ICLR_FlowMatching}, which learns a velocity field that transports Gaussian noise to image latents. Let $z_{\mathrm{data}}$ denote a normalized image latent and $\epsilon\sim\mathcal N(0,I)$ denote independent Gaussian noise. With flow time $\tau\in[0,1]$ increasing from noise to data, the linear interpolation path and its target velocity are
\begin{align}
  z_\tau&=(1-\tau)\epsilon+\tau z_{\mathrm{data}},
    \label{eq:appendix_flow_path}\\
  u_\tau&=\frac{\mathrm d z_\tau}{\mathrm d\tau}
    =z_{\mathrm{data}}-\epsilon.
    \label{eq:appendix_flow_target}
\end{align}
The conditional DiT $v_\theta(z_\tau,\tau,c)$ learns this velocity through the flow-matching objective
\begin{equation}
  \mathcal L_{\mathrm{FM}}
  =\mathbb E_{z_{\mathrm{data}},\epsilon,\tau,c}
    \!\left[\left\|v_\theta(z_\tau,\tau,c)-u_\tau\right\|_2^2\right],
  \label{eq:appendix_flow_loss}
\end{equation}
where $c$ denotes the text condition in the pretrained model and $\tau$ follows its training-time sampling distribution.

We generate latents by solving the flow ODE from a Gaussian initial state.
\begin{equation}
  \frac{\mathrm d z(\tau)}{\mathrm d\tau}
    =v_\theta(z(\tau),\tau,c),
  \qquad z(0)=\epsilon.
  \label{eq:appendix_flow_ode}
\end{equation}
A forward Euler step on an increasing time grid gives
\begin{equation}
  z^{k+1}=z^k+(\tau_{k+1}-\tau_k)
    v_\theta(z^k,\tau_k,c).
  \label{eq:appendix_flow_euler}
\end{equation}
With $\tau_0=0$ and $\tau_1=1$, a single Euler step gives $z^1=\epsilon+v_\theta(\epsilon,0,c)$, approximating the pretrained flow.

\paragraph{Application to PICS.}
We adapt this pretrained flow prior for source-conditioned reconstruction. The DiT receives decoded synthesis features $\hat l$ and visual conditions $\mathbf{C}_{\mathrm{vis}}=T(\hat y)$ derived from the bitstream. We parameterize the adapted predictor $\Phi_\theta$ by the scheduler timestep $t_0=1000$, with the internal conversion $\tau=1-t_0/1000=0$ supplying the backbone time embedding. The learned projection $P(\hat l)$ initializes the reconstruction state, and the adapted DiT predicts a correction $\Phi_\theta(\hat l,t_0,\mathbf{C}_{\mathrm{vis}})$ in the same expanded reconstruction space. The entropy-conditioned gate $g$ modulates the correction strength~\cite{Shi_26_CVPR_DiTIC}. This preserves the additive form of the Euler update while allowing the DiT input and reconstruction state to use different channel widths through the IPAs, as formalized in the Single-Step Reconstruction subsection of the main paper. The synthesis features, visual conditions, and gate are determined from the bitstream, yielding deterministic reconstruction. The objective $\mathcal L_{\mathrm{FM}}$ describes backbone pretraining, while PICS adapts the pretrained predictor with the compression objectives in the main paper.

\subsection{Information-Preserving Adaptors}
\label{sec:appendix_adaptors}
\begin{figure*}[t]
  \centering
  \includegraphics[width=\textwidth]{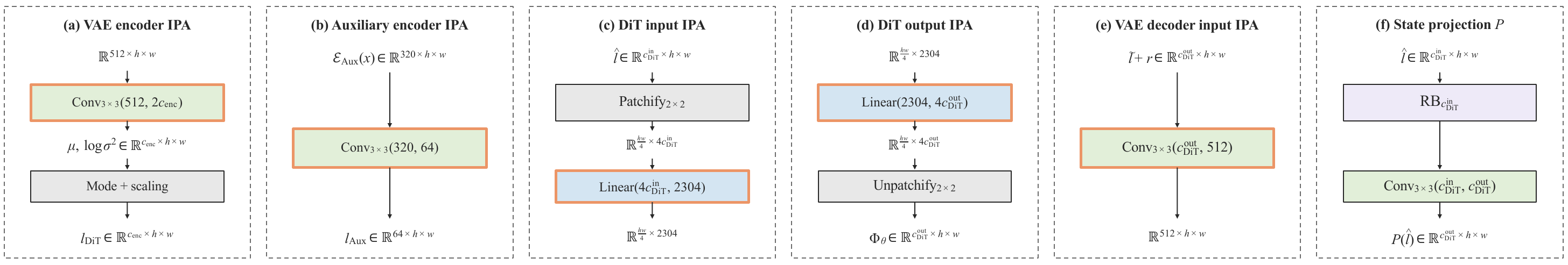}
  \caption{Architecture of the IPAs and reconstruction-state projection $P$, using the channel notation of the main paper. The VAE hidden widths at these projections remain 512, and the DiT hidden width remains 2304. DiT projections use $2\times2$ patches, with $4c_{\mathrm{DiT}}^{\mathrm{in}}=1280$ and $4c_{\mathrm{DiT}}^{\mathrm{out}}=512$. $\mathrm{RB}$ denotes residual feature refinement.}
  \label{fig:appendix_adaptors}
\end{figure*}

Our Information-Preserving Adaptors (IPAs) use the main paper's channel widths: $c_{\mathrm{enc}}=128$ for source encoding, $c_{\mathrm{DiT}}^{\mathrm{in}}=320$ for the DiT input, and $c_{\mathrm{DiT}}^{\mathrm{out}}=128$ for reconstruction. The DiT input IPA processes $\hat l$, while the DiT output IPA predicts the correction used to form $\tilde l$. The VAE decoder input IPA receives $\tilde l+r$. Figure~\ref{fig:appendix_adaptors} summarizes these mappings and the state projection $P$.

\subsection{Visual Conditioning}
\label{sec:appendix_visual_tokens}
The token extractor applies residual refinement and channel projection to $\hat y$, followed by adaptive average pooling to $16\times16$ and LayerNorm. The resulting 256 visual tokens, each with 2304 dimensions, replace text embeddings in the DiT's native conditioning pathway and interact with image patch tokens through joint attention. This provides source-dependent guidance directly from the coded representation, with no additional side information.

\subsection{Training Configuration}
\label{sec:appendix_lora}
We use rank-8 LoRA for the VAE encoder, DiT, and VAE decoder, with $\alpha=8$ and zero dropout. LoRA adapts convolutional and attention layers in the VAE, and attention and adaptive-normalization projections in the DiT. The fixed low rank keeps backbone adaptation compact, placing the emphasis on IPA design and deferred decoder adaptation. LoRA uses Gaussian initialization with zero-initialized output weights; IPAs are optimized according to the stage schedule in the main paper.

\label{sec:appendix_losses}
The reconstruction weights are $(\gamma_1,\gamma_2,\gamma_3,\gamma_4)=(2,1,0.1,1)$ for MSE, LPIPS, CLIP loss, and EA-DISTS~\cite{Li_25_NeurIPS_D3SR,Zhang_26_CVPR_AEIC}, respectively. Our discriminator follows the StableCodec architecture~\cite{Zhang_25_ICCV_StableCodec}. We weight the adversarial loss by $\beta=0.1$ in Stage~3.

\section{Visualization and Evaluation Protocols}
\label{sec:appendix_protocols}

\subsection{Latent Visualization Protocol}
\label{sec:appendix_latent_visualization}
Figure~1 of the main paper compares VAE encoder outputs and decoder inputs after generative reconstruction, showing how spatial structure is preserved at both interfaces.

\paragraph{Spatial heatmaps.}
Using the same Kodak image (\texttt{kodim23}), we subtract each channel's spatial mean and divide each latent tensor by its global centered root-mean-square (RMS) value. We then compute the RMS across channels at each spatial position. All maps use a shared color scale and nearest-neighbor enlargement from their native grids. This protocol supports comparisons of spatial activation patterns across resolutions and channel counts.

\paragraph{Latent distributions.}
Each point represents the full channel vector at one spatial position. We pool features from all 24 Kodak images, sampling 1,024 positions per image for PICS, StableCodec, and OneDC and using all 384 native positions for DiT-IC. For each method and interface, we subtract the pooled per-channel means and normalize by a single RMS vector norm. We fit a separate two-dimensional t-SNE embedding with identical settings, a fixed random seed, and PCA initialization. Each embedding is centered, scaled isotropically to unit RMS radius, and plotted with shared axis limits. Colors encode Gaussian kernel density in the displayed space, using a common bandwidth and color scale. Separate embeddings retain native channel counts, while shared preprocessing and rendering support qualitative comparisons of local feature organization.

\subsection{FID Evaluation Protocol}
\label{sec:appendix_fid}
We evaluate all methods using the same patch extraction, feature extractor, and aggregation protocol on each dataset.

\paragraph{Patch extraction.}
Reference and reconstructed images are converted to RGB and paired by filename. Each pair is cropped to its common top-left extent if the dimensions differ. We extract $256\times256$ patches on two grids with a stride of 256 pixels in both directions, starting at $(0,0)$ and $(128,128)$, respectively. The shifted grid is used when both image dimensions are at least 384 pixels. We retain fully contained patches, discarding incomplete boundary patches without padding or resizing the source images. Reference and reconstruction patches use identical coordinates.

\paragraph{FID computation.}
We use TorchMetrics' \texttt{FrechetInceptionDistance} with its default Inception-v3~\cite{Szegedy_2016_CVPR_InceptionV3} extractor and 2,048-dimensional features. RGB patches are passed as 8-bit integers in $[0,255]$ in batches of eight; the extractor applies its standard $299\times299$ resizing. For each dataset and rate operating point, we accumulate reference and reconstruction feature statistics over all patches and compute one FID from the two pooled means and covariances. Every patch contributes equally to these dataset-level statistics.

\paragraph{Dataset selection.}
We report FID on DIV2K and CLIC 2020, whose larger image collections provide more samples for distribution-level evaluation. Under this protocol, Kodak's 24 images yield eight patches each, totaling 192 patches per distribution. This sample count is small relative to the 2,048-dimensional feature space, and overlapping crops share image content. We therefore use Kodak for paired fidelity and perceptual similarity evaluation, and assess distribution-level realism on the larger datasets.

\subsection{BD-rate Computation}
\label{sec:appendix_bd_rate}

For each method $m$, let $\{(r_{m,i},q_{m,i})\}_{i=1}^{n_m}$
denote its measured rate--quality points, where $r_{m,i}>0$ is
the bitrate in bpp and $q_{m,i}$ is the quality metric. We sort
the points by increasing $q_{m,i}$ and construct
\begin{equation}
  f_m(q)=\operatorname{PCHIP}\!\left(
    \{(q_{m,i},\ln r_{m,i})\}_{i=1}^{n_m}\right)(q).
  \label{eq:appendix_pchip}
\end{equation}
PCHIP is a piecewise cubic Hermite interpolant that preserves
the monotonicity of monotone input data. This avoids the
non-monotone overshoot that can arise from a global cubic fit
between sparsely sampled operating points. All measured points
define each interpolant, including those outside the integration
interval. We use LPIPS and DISTS directly, without negation or a log transform. For MS-SSIM, we use the original $[0,1]$ scale.

For an anchor curve $a$ and a test curve $t$, the common quality
interval is $[q_{\mathrm{lo}},q_{\mathrm{hi}}]$, with
\begin{equation}
  \begin{aligned}
    q_{\mathrm{lo}}&=\max\!\left(\min_i q_{a,i},\min_j q_{t,j}\right),\\
    q_{\mathrm{hi}}&=\min\!\left(\max_i q_{a,i},\max_j q_{t,j}\right).
  \end{aligned}
  \label{eq:appendix_overlap}
\end{equation}
We require $q_{\mathrm{hi}}>q_{\mathrm{lo}}$ and do not extrapolate
either curve. The average log-rate difference and BD-rate are
\begin{align}
  \Delta_{t|a}
    &=\frac{\displaystyle\int_{q_{\mathrm{lo}}}^{q_{\mathrm{hi}}}
       \bigl[f_t(q)-f_a(q)\bigr]\,\mathrm{d}q}
       {q_{\mathrm{hi}}-q_{\mathrm{lo}}},
       \label{eq:appendix_log_rate_difference}\\
  \operatorname{BD\text{-}rate}_{t|a}
    &=100\bigl[\exp(\Delta_{t|a})-1\bigr]\%.
       \label{eq:appendix_bd_rate}
\end{align}
We integrate the piecewise polynomials exactly using their
antiderivatives, rather than sampling the fitted curves.
Our implementation uses SciPy's \texttt{PchipInterpolator}
with extrapolation disabled and its \texttt{integrate} method.
Negative BD-rate indicates that the test method requires fewer
bits at matched quality; positive BD-rate indicates more bits.
This sign convention also holds for lower-is-better metrics.
Averaging in log space gives the geometric mean of the rate ratios over the common quality interval.

\section{Runtime at Higher Resolution}
\label{sec:appendix_runtime}
Table~\ref{tab:appendix_runtime_1024} reports mean encoding and decoding latency for $1024\times1024$ images on a single NVIDIA GeForce RTX 5090 GPU. PICS reduces total coding latency by 26.51\% relative to StableCodec, from 833.92\,ms to 612.88\,ms, supporting efficient high-resolution reconstruction.

\begin{table}[!h]
  \centering
  \small
  \caption{Mean encoding and decoding latency (ms) for $1024{\times}1024$ images on one RTX 5090. Lower is better.}
  \label{tab:appendix_runtime_1024}
  \setlength{\tabcolsep}{6pt}
  \renewcommand{\arraystretch}{1.12}
  \sisetup{detect-weight=true, mode=text}
  \begin{tabular}{@{}l
                  S[table-format=3.2]
                  S[table-format=5.2]@{}}
    \toprule
    \multicolumn{1}{l}{Method} & {Enc. (ms)$\downarrow$} & {Dec. (ms)$\downarrow$} \\
    \midrule
    MS-ILLM     & 400.77 &   362.09 \\
    GLC         &  94.21 &   160.88 \\
    DLF         & 144.18 &   320.77 \\
    DiffEIC     & 320.42 & 14772.84 \\
    OSCAR       & 163.46 &   498.20 \\
    StableCodec & 294.90 &   539.02 \\
    AEIC-SE     &  13.38 &   199.48 \\
    DiT-IC      & 235.42 &   368.71 \\
    OneDC       & 170.55 &   378.56 \\
    \midrule
    \rowcolor{gray!12}
    \textbf{Ours} & 172.21 & 440.67 \\
    \bottomrule
  \end{tabular}
\end{table}

\section{Comprehensive Rate--Quality Comparison}
\label{sec:appendix_bd_comparison}
PICS achieves the best average BD-rates in LPIPS, DISTS, and PSNR among the methods with comparable measurements in Table~\ref{tab:appendix_bd_comparison}. Relative to StableCodec~\cite{Zhang_25_ICCV_StableCodec}, PICS reduces bitrate by 52.64\% at matched LPIPS and 39.61\% at matched DISTS, averaged across Kodak, DIV2K, and CLIC. On DIV2K and CLIC, the savings exceed 50\% under both perceptual metrics: 55.42\% and 54.49\% for LPIPS, and 50.56\% and 50.34\% for DISTS, respectively. These perceptual gains accompany an average bitrate reduction of 20.45\% at matched PSNR, demonstrating improved coding efficiency across both perceptual quality and pixel fidelity.

Direct comparisons with PICS as the anchor in Table~\ref{tab:appendix_bd_comparison_ours} further quantify the advantage over recent codecs. At matched DISTS, NeFIC~\cite{Chen_26_ECCV_NeFIC} requires 24.99\% more bitrate than PICS on average across the three datasets, with the overhead reaching 38.04\% on DIV2K and 37.36\% on CLIC. AEIC-ME~\cite{Zhang_26_CVPR_AEIC} and DiT-IC~\cite{Shi_26_CVPR_DiTIC} require 28.95\% and 46.63\% more bitrate on average at matched DISTS, respectively. At matched LPIPS, AEIC-ME and DiT-IC require 82.09\% and 28.98\% more bitrate than PICS on average, respectively. Together, these results establish PICS's advantage in reducing the coding cost of perceptually faithful reconstruction across the evaluated benchmarks.

\newcommand{\best}[1]{\textcolor{bdbest}{$\mathbf{#1}$}}
\newcommand{\bdsecondvalue}[1]{\textcolor{bdsecond}{$\mathbf{#1}$}}
\newlength{\bdmethodwidth}
\newlength{\bdnumericwidth}

\section{Baseline Implementations and Evaluation}
\label{sec:appendix_baseline_code}
We run the baseline codecs ourselves using their official implementations and evaluate the resulting reconstructions under the experimental setup described in the main paper. Table~\ref{tab:appendix_baseline_code} documents the source repositories used in these experiments, grouped by codec family. AEIC-SE and AEIC-ME share the AEIC implementation.

\section{Additional Visual Comparisons}
\label{sec:appendix_visual}
\begin{figure*}[p]
\centering
\includegraphics[width=\textwidth,height=\dimexpr\textheight-4\baselineskip\relax,keepaspectratio]{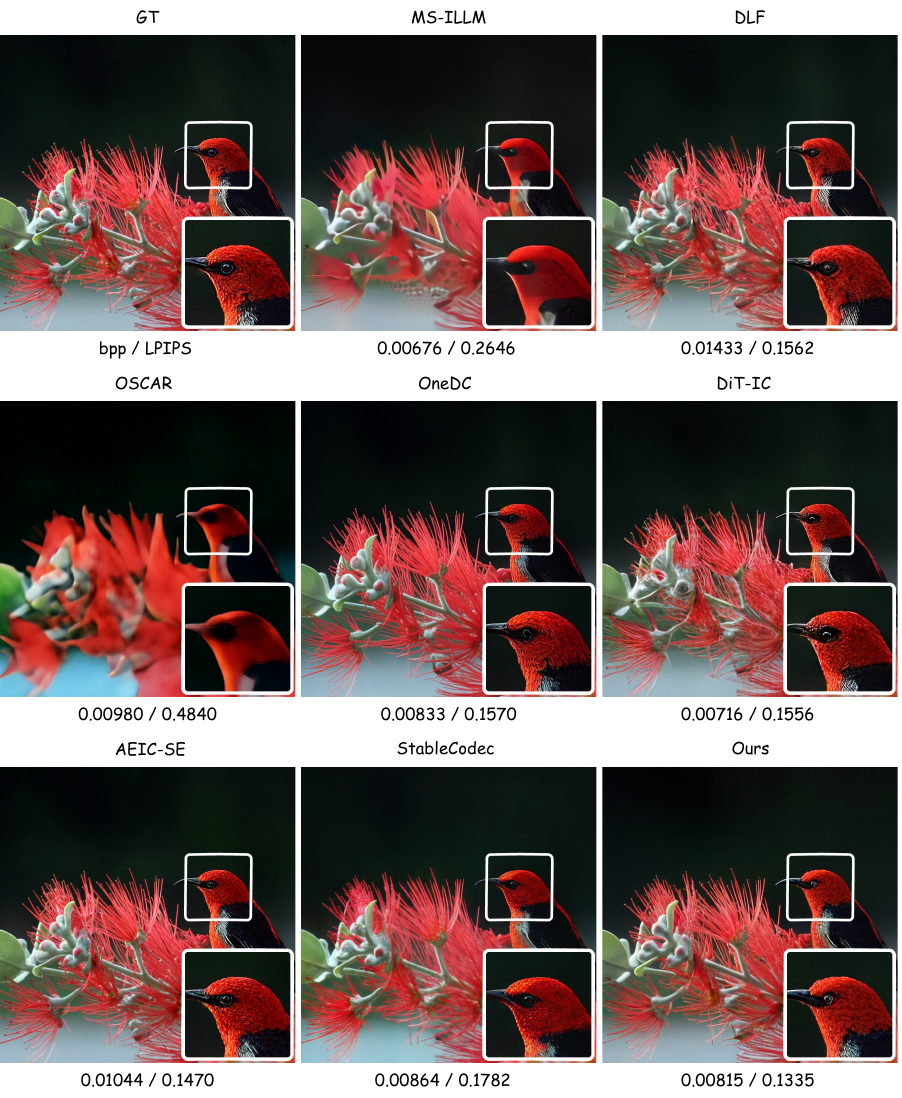}
\caption{Additional visual comparison at ultra-low bitrates (example 1 of 6). Insets magnify the marked regions. Labels report full-image bpp / LPIPS (lower is better); crops are for visualization. GT denotes the original image; Ours denotes PICS.}
\label{fig:appendix_visual_1}
\end{figure*}

\begin{figure*}[p]
\centering
\includegraphics[width=\textwidth,height=\dimexpr\textheight-4\baselineskip\relax,keepaspectratio]{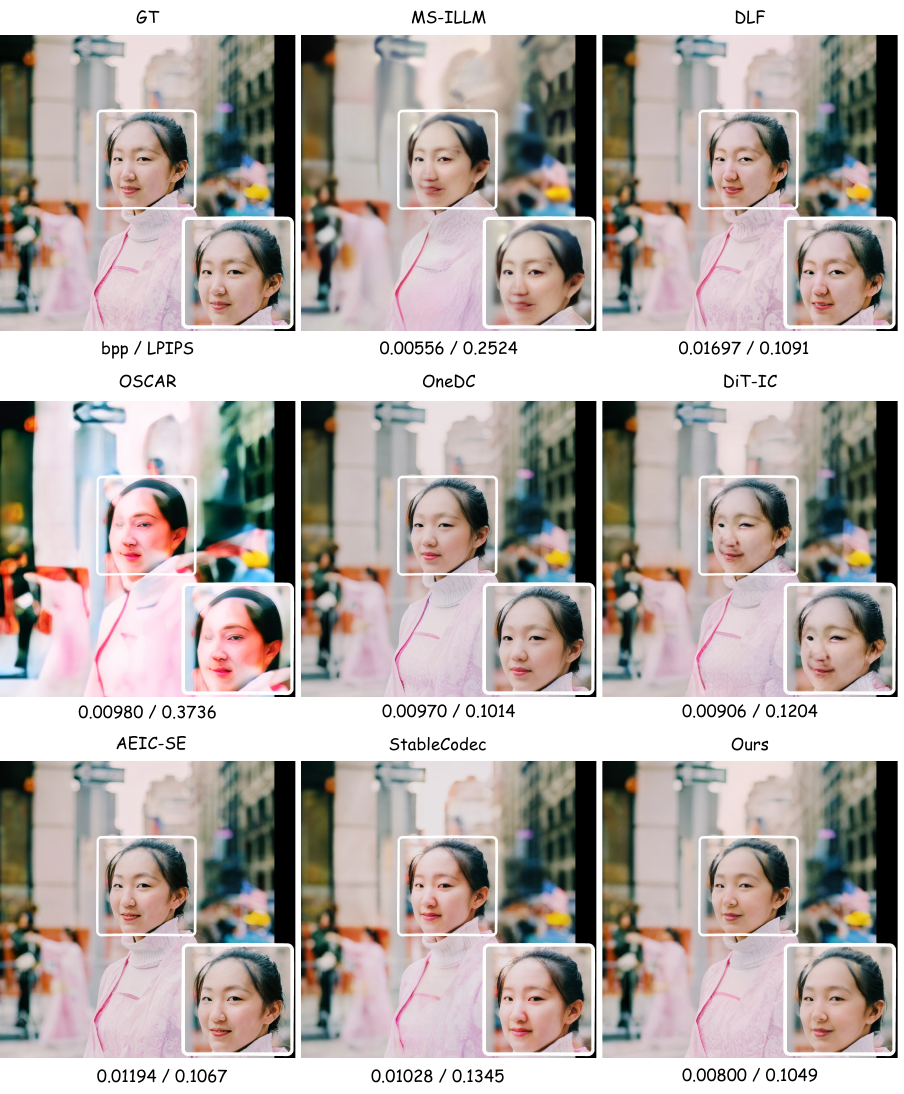}
\caption{Additional visual comparison at ultra-low bitrates (example 2 of 6). Insets magnify the marked regions. Labels report full-image bpp / LPIPS (lower is better); crops are for visualization. GT denotes the original image; Ours denotes PICS.}
\label{fig:appendix_visual_2}
\end{figure*}

\begin{figure*}[p]
\centering
\includegraphics[width=\textwidth,height=\dimexpr\textheight-4\baselineskip\relax,keepaspectratio]{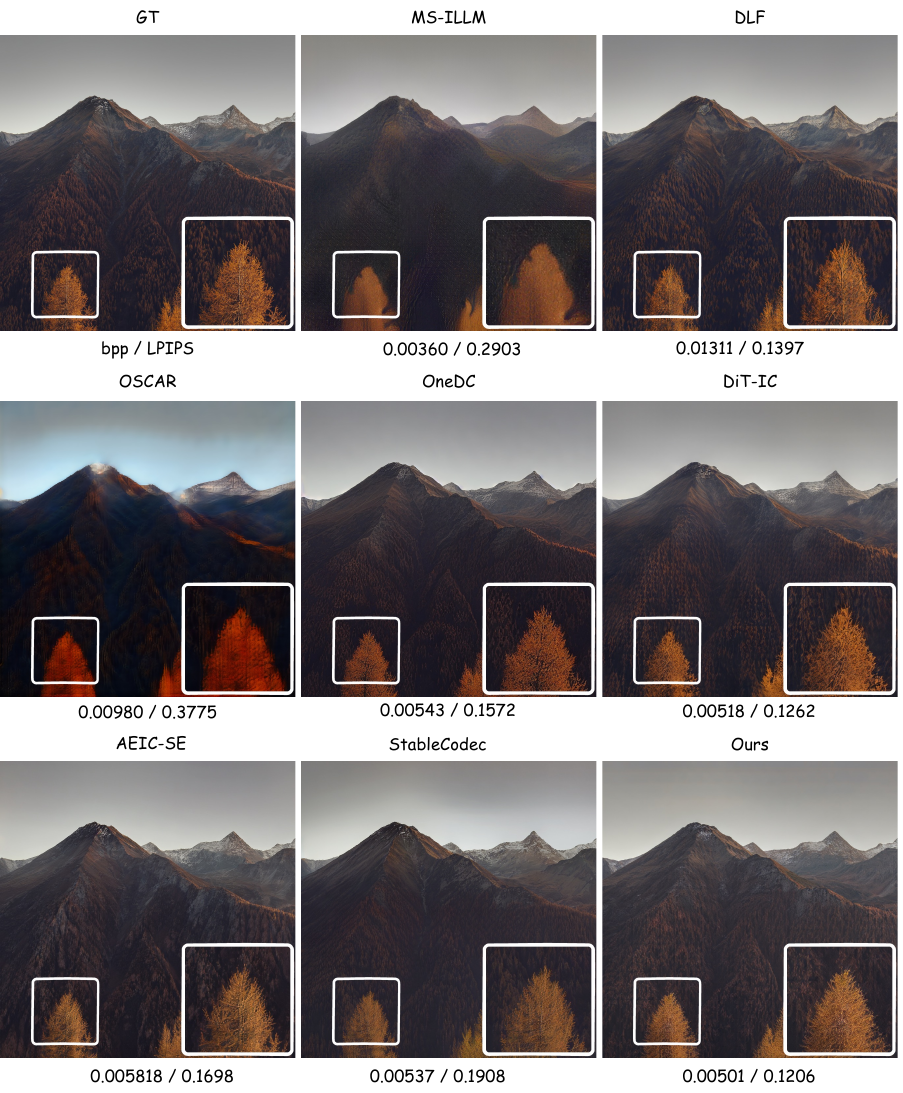}
\caption{Additional visual comparison at ultra-low bitrates (example 3 of 6). Insets magnify the marked regions. Labels report full-image bpp / LPIPS (lower is better); crops are for visualization. GT denotes the original image; Ours denotes PICS.}
\label{fig:appendix_visual_3}
\end{figure*}

\begin{figure*}[p]
\centering
\includegraphics[width=\textwidth,height=\dimexpr\textheight-4\baselineskip\relax,keepaspectratio]{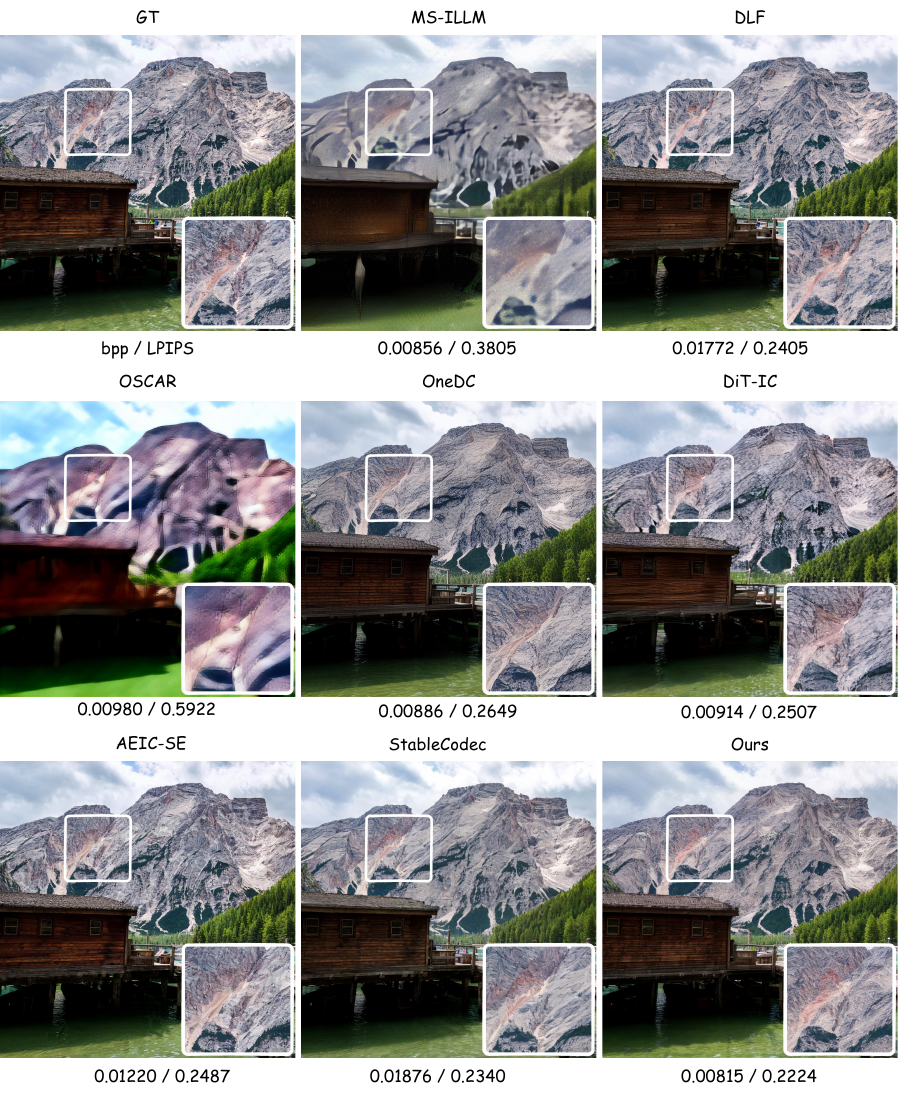}
\caption{Additional visual comparison at ultra-low bitrates (example 4 of 6). Insets magnify the marked regions. Labels report full-image bpp / LPIPS (lower is better); crops are for visualization. GT denotes the original image; Ours denotes PICS.}
\label{fig:appendix_visual_4}
\end{figure*}

\begin{figure*}[p]
\centering
\includegraphics[width=\textwidth,height=\dimexpr\textheight-4\baselineskip\relax,keepaspectratio]{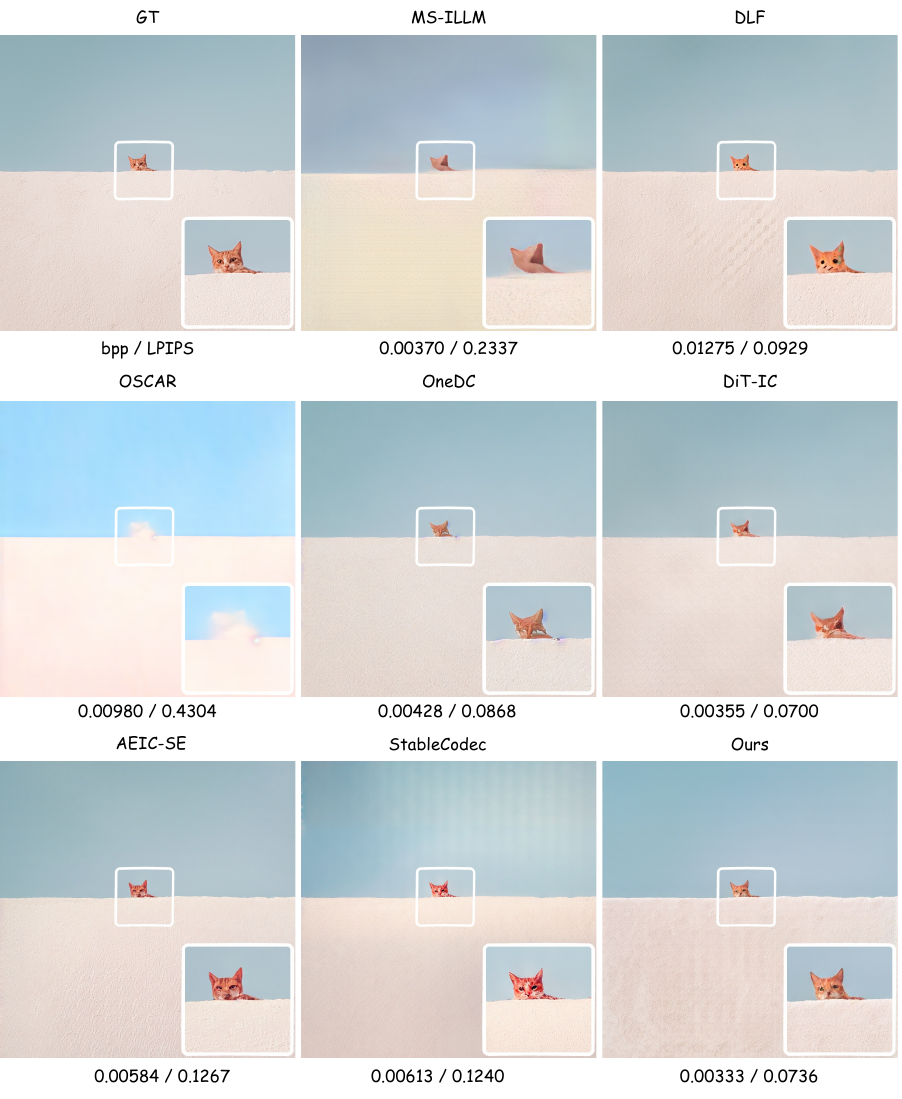}
\caption{Additional visual comparison at ultra-low bitrates (example 5 of 6). Insets magnify the marked regions. Labels report full-image bpp / LPIPS (lower is better); crops are for visualization. GT denotes the original image; Ours denotes PICS.}
\label{fig:appendix_visual_5}
\end{figure*}

\begin{figure*}[p]
\centering
\includegraphics[width=\textwidth,height=\dimexpr\textheight-4\baselineskip\relax,keepaspectratio]{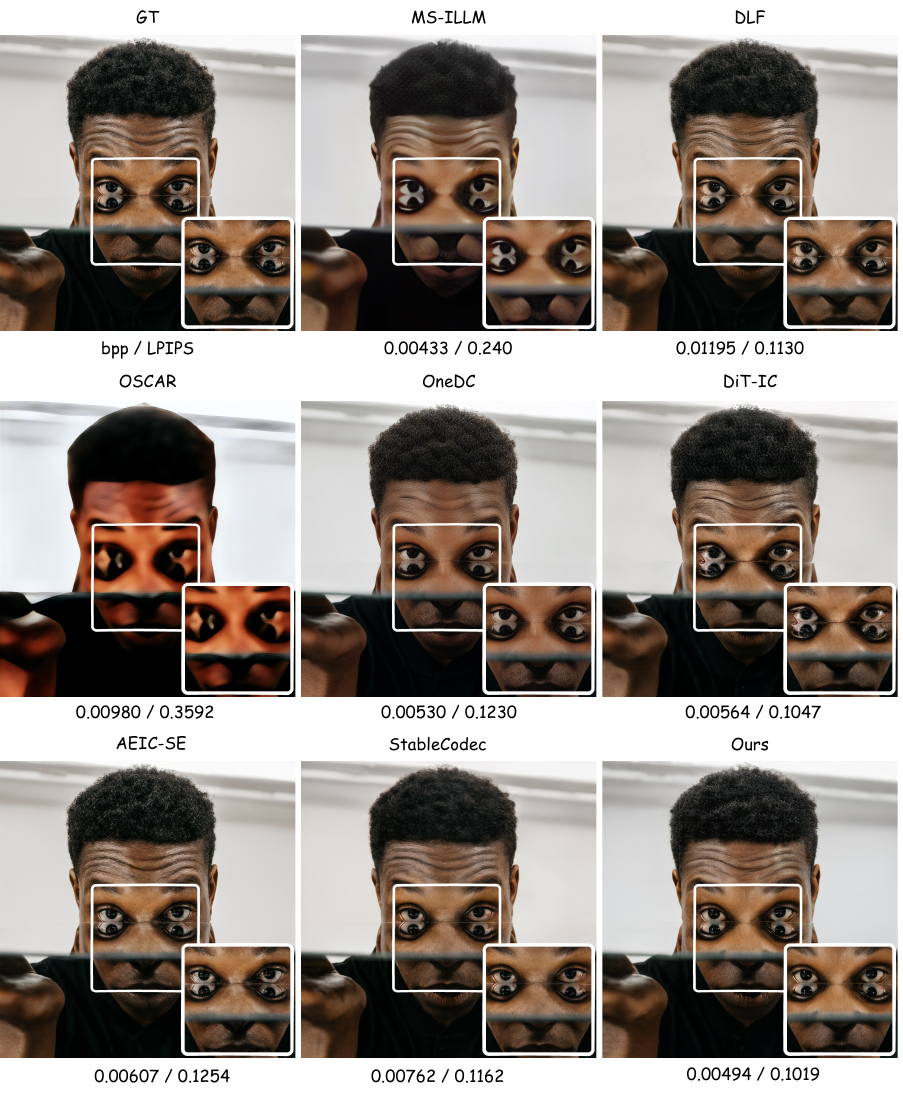}
\caption{Additional visual comparison at ultra-low bitrates (example 6 of 6). Insets magnify the marked regions. Labels report full-image bpp / LPIPS (lower is better); crops are for visualization. GT denotes the original image; Ours denotes PICS.}
\label{fig:appendix_visual_6}
\end{figure*}

\noindent
Figures~\ref{fig:appendix_visual_1}--\ref{fig:appendix_visual_6} compare PICS with MS-ILLM~\cite{Muckley_2023_ICML_MSILLM}, DLF~\cite{Xue_25_ICCV_DLF}, OSCAR~\cite{Guo_25_NeurIPS_oscar}, OneDC~\cite{Xue_25_NeurIPS_OneDC}, DiT-IC~\cite{Shi_26_CVPR_DiTIC}, AEIC-SE~\cite{Zhang_26_CVPR_AEIC}, and StableCodec~\cite{Zhang_25_ICCV_StableCodec} at ultra-low bitrates. Magnified regions highlight local reconstruction detail, while the bpp / LPIPS labels report the rate--quality operating point of each method. In Figure~\ref{fig:appendix_visual_1}, PICS achieves the lowest LPIPS among the displayed reconstructions, reaching 0.1335 at 0.00815 bpp.

\raggedbottom
\clearpage
\begin{table*}[p]
\centering
\begin{minipage}{\textwidth}
  \centering
  \footnotesize
  \caption{Comprehensive BD-rate (\%) comparison relative to StableCodec. All entries are BD-rates, so lower is better (including PSNR); negative values indicate bitrate savings. The \textcolor{bdbest}{\textbf{best}} and \textcolor{bdsecond}{\textbf{second-best}} results among the displayed methods in each column are highlighted. N/A denotes missing measurements or results obtained under incompatible measurement settings. An average is reported only when comparable measurements are available for all three datasets.}
  \label{tab:appendix_bd_comparison}
  \setlength{\tabcolsep}{2.5pt}
  \renewcommand{\arraystretch}{1.12}
  \settowidth{\bdmethodwidth}{StableCodec (ICCV'25)~\cite{Zhang_25_ICCV_StableCodec}}
  \setlength{\bdnumericwidth}{\dimexpr(\textwidth-\bdmethodwidth-24\tabcolsep)/12\relax}
  \sisetup{table-format=-3.2, table-column-width=\bdnumericwidth, table-text-alignment=right, detect-weight=true, mode=text}
  \begin{tabular}{@{}l *{12}{S}@{}}
    \toprule
    \multirow{2}{*}{Method}
      & \multicolumn{3}{c}{LPIPS$\downarrow$}
      & \multicolumn{3}{c}{DISTS$\downarrow$}
      & \multicolumn{3}{c}{PSNR$\downarrow$}
      & \multicolumn{3}{c}{Average$\downarrow$} \\
    \cmidrule(lr){2-4}\cmidrule(lr){5-7}\cmidrule(lr){8-10}\cmidrule(l){11-13}
      & {Kodak} & {DIV2K} & {CLIC}
      & {Kodak} & {DIV2K} & {CLIC}
      & {Kodak} & {DIV2K} & {CLIC}
      & {LPIPS} & {DISTS} & {PSNR} \\
    \midrule
    GLC (CVPR'24)~\cite{Jia_24_CVPR_GLC}
      & 9.02 & -8.48 & -7.43
      & 57.16 & 37.33 & 22.32
      & 68.31 & 36.53 & 16.19
      & -2.30 & 38.94 & 40.35 \\
    DLF (ICCV'25)~\cite{Xue_25_ICCV_DLF}
      & 13.58 & 19.17 & 12.91
      & 37.54 & 65.68 & 33.22
      & 67.81 & 79.81 & 70.42
      & 15.22 & 45.48 & 72.68 \\
    DiffEIC (TCSVT'25)~\cite{Li_25_TCSVT_DiffEIC}
      & 193.32 & 206.24 & 243.35
      & 261.26 & 321.24 & 361.26
      & 174.93 & 205.19 & 246.35
      & 214.30 & 314.59 & 208.82 \\
    OSCAR (NeurIPS'25)~\cite{Guo_25_NeurIPS_oscar}
      & 221.63 & 256.38 & 360.29
      & 232.88 & 344.71 & 432.49
      & 244.62 & 266.14 & 424.50
      & 279.44 & 336.69 & 311.75 \\
    \rowcolor{gray!6}
    StableCodec (ICCV'25)~\cite{Zhang_25_ICCV_StableCodec}
      & 0.00 & 0.00 & 0.00
      & 0.00 & 0.00 & 0.00
      & 0.00 & 0.00 & 0.00
      & 0.00 & 0.00 & 0.00 \\
    OneDC (NeurIPS'25)~\cite{Xue_25_NeurIPS_OneDC}
      & -34.82 & -40.48 & -39.81
      & 3.98 & -7.38 & -20.35
      & 39.32 & 12.95 & 4.84
      & -38.37 & -7.91 & 19.04 \\
    AEIC-SE (CVPR'26)~\cite{Zhang_26_CVPR_AEIC}
      & -5.76 & -20.34 & -18.62
      & -13.00 & {\bdsecondvalue{-33.46}} & {\bdsecondvalue{-31.87}}
      & 17.76 & 4.82 & -11.65
      & -14.91 & -26.11 & 3.64 \\
    AEIC-ME (CVPR'26)~\cite{Zhang_26_CVPR_AEIC}
      & -9.08 & -12.80 & -8.72
      & {\bdsecondvalue{-17.77}} & -31.51 & -30.84
      & {\bdsecondvalue{-11.04}} & {\bdsecondvalue{-14.59}} & {\bdsecondvalue{-24.81}}
      & -10.20 & {\bdsecondvalue{-26.71}} & {\bdsecondvalue{-16.81}} \\
    DiT-IC (CVPR'26)~\cite{Shi_26_CVPR_DiTIC}
      & -33.96 & -43.76 & {\bdsecondvalue{-51.64}}
      & 7.60 & -14.42 & -23.08
      & 54.38 & 34.90 & 11.62
      & -43.12 & -9.97 & 33.63 \\
    NeFIC (ECCV'26)~\cite{Chen_26_ECCV_NeFIC}
      & {\bdsecondvalue{-45.95}} & {\bdsecondvalue{-49.12}} & -48.21
      & -17.76 & -24.33 & -23.78
      & {N/A} & {N/A} & {N/A}
      & {\bdsecondvalue{-47.76}} & -21.96 & {N/A} \\
    \midrule
    \rowcolor{gray!12}
    \textbf{PICS (Ours)}
      & {\best{-48.00}} & {\best{-55.42}} & {\best{-54.49}}
      & {\best{-17.93}} & {\best{-50.56}} & {\best{-50.34}}
      & {\best{-15.76}} & {\best{-20.55}} & {\best{-25.05}}
      & {\best{-52.64}} & {\best{-39.61}} & {\best{-20.45}} \\
    \bottomrule
  \end{tabular}
\end{minipage}
\par\vspace{12pt}
\begin{minipage}{\textwidth}
  \centering
  \footnotesize
  \caption{Comprehensive BD-rate (\%) comparison relative to PICS (Ours). All entries are BD-rates, so lower is better (including PSNR); negative values indicate bitrate savings. The \textcolor{bdbest}{\textbf{best}} and \textcolor{bdsecond}{\textbf{second-best}} results among the displayed methods in each column are highlighted. N/A denotes missing measurements or results obtained under incompatible measurement settings. An average is reported only when comparable measurements are available for all three datasets.}
  \label{tab:appendix_bd_comparison_ours}
  \setlength{\tabcolsep}{2.5pt}
  \renewcommand{\arraystretch}{1.12}
  \settowidth{\bdmethodwidth}{StableCodec (ICCV'25)~\cite{Zhang_25_ICCV_StableCodec}}
  \setlength{\bdnumericwidth}{\dimexpr(\textwidth-\bdmethodwidth-24\tabcolsep)/12\relax}
  \sisetup{table-format=-3.2, table-column-width=\bdnumericwidth, table-text-alignment=right, detect-weight=true, mode=text}
  \begin{tabular}{@{}l *{12}{S}@{}}
    \toprule
    \multirow{2}{*}{Method}
      & \multicolumn{3}{c}{LPIPS$\downarrow$}
      & \multicolumn{3}{c}{DISTS$\downarrow$}
      & \multicolumn{3}{c}{PSNR$\downarrow$}
      & \multicolumn{3}{c}{Average$\downarrow$} \\
    \cmidrule(lr){2-4}\cmidrule(lr){5-7}\cmidrule(lr){8-10}\cmidrule(l){11-13}
      & {Kodak} & {DIV2K} & {CLIC}
      & {Kodak} & {DIV2K} & {CLIC}
      & {Kodak} & {DIV2K} & {CLIC}
      & {LPIPS} & {DISTS} & {PSNR} \\
    \midrule
    GLC (CVPR'24)~\cite{Jia_24_CVPR_GLC}
      & 123.24 & 129.11 & 139.47
      & 98.06 & 186.59 & 167.77
      & 96.35 & 85.20 & 98.54
      & 130.61 & 150.81 & 93.36 \\
    DLF (ICCV'25)~\cite{Xue_25_ICCV_DLF}
      & 115.66 & 153.95 & 128.27
      & 59.18 & 198.76 & 151.76
      & 103.99 & 108.32 & 96.10
      & 132.63 & 136.56 & 102.80 \\
    DiffEIC (TCSVT'25)~\cite{Li_25_TCSVT_DiffEIC}
      & 372.47 & 406.14 & 460.88
      & 322.00 & 501.46 & 557.97
      & 219.49 & 269.66 & 368.00
      & 413.16 & 460.47 & 285.72 \\
    OSCAR (NeurIPS'25)~\cite{Guo_25_NeurIPS_oscar}
      & 471.14 & 592.69 & 773.65
      & 283.96 & 612.31 & 720.92
      & 294.68 & 332.84 & 528.94
      & 612.49 & 539.07 & 385.49 \\
    StableCodec (ICCV'25)~\cite{Zhang_25_ICCV_StableCodec}
      & 92.32 & 124.32 & 119.72
      & 21.84 & 102.27 & 101.35
      & 18.71 & 25.86 & 33.42
      & 112.12 & 75.15 & 26.00 \\
    OneDC (NeurIPS'25)~\cite{Xue_25_NeurIPS_OneDC}
      & 38.27 & 41.50 & 37.32
      & 28.10 & 90.13 & 65.25
      & 66.53 & 43.11 & 41.78
      & 39.03 & 61.16 & 50.47 \\
    AEIC-SE (CVPR'26)~\cite{Zhang_26_CVPR_AEIC}
      & 82.27 & 74.65 & 71.62
      & 5.52 & 42.20 & 39.57
      & 40.08 & 32.71 & 18.93
      & 76.18 & 29.10 & 30.57 \\
    AEIC-ME (CVPR'26)~\cite{Zhang_26_CVPR_AEIC}
      & 75.54 & 85.31 & 85.43
      & 2.32 & 43.84 & 40.68
      & {\bdsecondvalue{5.93}} & {\bdsecondvalue{8.97}} & {\bdsecondvalue{4.27}}
      & 82.09 & 28.95 & {\bdsecondvalue{6.39}} \\
    DiT-IC (CVPR'26)~\cite{Shi_26_CVPR_DiTIC}
      & 31.11 & 34.61 & 21.24
      & 27.79 & 63.33 & 48.77
      & 81.78 & 67.32 & 47.05
      & 28.98 & 46.63 & 65.39 \\
    NeFIC (ECCV'26)~\cite{Chen_26_ECCV_NeFIC}
      & {\bdsecondvalue{1.20}} & {\bdsecondvalue{6.39}} & {\bdsecondvalue{4.40}}
      & {\best{-0.44}} & {\bdsecondvalue{38.04}} & {\bdsecondvalue{37.36}}
      & {N/A} & {N/A} & {N/A}
      & {\bdsecondvalue{4.00}} & {\bdsecondvalue{24.99}} & {N/A} \\
    \midrule
    \rowcolor{gray!12}
    \textbf{PICS (Ours)}
      & {\best{0.00}} & {\best{0.00}} & {\best{0.00}}
      & {\bdsecondvalue{0.00}} & {\best{0.00}} & {\best{0.00}}
      & {\best{0.00}} & {\best{0.00}} & {\best{0.00}}
      & {\best{0.00}} & {\best{0.00}} & {\best{0.00}} \\
    \bottomrule
  \end{tabular}
\end{minipage}
\par\vspace{12pt}
\begin{minipage}{\textwidth}
  \centering
  \small
  \setlength{\tabcolsep}{5pt}
  \renewcommand{\arraystretch}{1.1}
  \urlstyle{same}
  \newcommand{\baselinerepo}[2]{\href{#1}{\textcolor{black}{\faGithub}\enspace\nolinkurl{#2}}}
  \newsavebox{\baselinecodetable}
  \begin{lrbox}{\baselinecodetable}
  \begin{tabular}{@{}c c l c l@{}}
    \toprule
    \textbf{Category} & \textbf{\#} & \textbf{Method} & \textbf{Venue} & \textbf{Code repository} \\
    \midrule
    \rowcolor{gray!6}
    \cellcolor{white}GAN-based & 1 & MS-ILLM~\cite{Muckley_2023_ICML_MSILLM} & ICML'23
      & \baselinerepo{https://github.com/facebookresearch/NeuralCompression/tree/main/projects/illm}{facebookresearch/NeuralCompression} \\
    \midrule
    \cellcolor{white} & 2 & GLC~\cite{Jia_24_CVPR_GLC} & CVPR'24
      & \baselinerepo{https://github.com/jzyustc/GLC}{jzyustc/GLC} \\
    \rowcolor{gray!6}
    \cellcolor{white}\multirow{-2}{*}{Tokenizer-based} & 3 & DLF~\cite{Xue_25_ICCV_DLF} & ICCV'25
      & \baselinerepo{https://github.com/dlfcodec/Dual-generative-Latent-Fusion}{dlfcodec/Dual-generative-Latent-Fusion} \\
    \midrule
    \cellcolor{white} & 4 & DiffEIC~\cite{Li_25_TCSVT_DiffEIC} & TCSVT'25
      & \baselinerepo{https://github.com/huai-chang/DiffEIC}{huai-chang/DiffEIC} \\
    \rowcolor{gray!6}
    \cellcolor{white} & 5 & OSCAR~\cite{Guo_25_NeurIPS_oscar} & NeurIPS'25
      & \baselinerepo{https://github.com/jp-guo/OSCAR}{jp-guo/OSCAR} \\
    \cellcolor{white} & 6 & StableCodec~\cite{Zhang_25_ICCV_StableCodec} & ICCV'25
      & \baselinerepo{https://github.com/LuizScarlet/StableCodec}{LuizScarlet/StableCodec} \\
    \rowcolor{gray!6}
    \cellcolor{white} & 7 & AEIC-SE / ME~\cite{Zhang_26_CVPR_AEIC} & CVPR'26
      & \baselinerepo{https://github.com/LuizScarlet/AEIC}{LuizScarlet/AEIC} \\
    \cellcolor{white} & 8 & DiT-IC~\cite{Shi_26_CVPR_DiTIC} & CVPR'26
      & \baselinerepo{https://github.com/Eric-qi/DiT-IC}{Eric-qi/DiT-IC} \\
    \rowcolor{gray!6}
    \cellcolor{white} & 9 & OneDC~\cite{Xue_25_NeurIPS_OneDC} & NeurIPS'25
      & \baselinerepo{https://github.com/onedc-codec/onedc}{onedc-codec/onedc} \\
    \cellcolor{white}\multirow{-7}{*}{Diffusion-based} & 10 & NeFIC~\cite{Chen_26_ECCV_NeFIC} & ECCV'26
      & \baselinerepo{https://github.com/UnoC-727/NeFIC}{UnoC-727/NeFIC} \\
    \bottomrule
  \end{tabular}
  \end{lrbox}
  \captionsetup{font=small,width=\textwidth}
  \caption{Official implementations used in our baseline evaluations, grouped by codec family. Repository links are clickable.}
  \label{tab:appendix_baseline_code}
  \usebox{\baselinecodetable}
\end{minipage}
\end{table*}

\clearpage

\clearpage
\end{document}